\documentclass[11pt]{article}
\PassOptionsToPackage{table}{xcolor}
\usepackage[preprint]{acl}
\usepackage{times,latexsym}
\usepackage[T1]{fontenc}
\usepackage[utf8]{inputenc}
\usepackage{microtype,inconsolata}
\usepackage{amsmath,amssymb,amsthm,mathtools}
\usepackage{booktabs,array,tabularx,multirow}
\usepackage{graphicx}
\usepackage{tcolorbox}
\tcbuselibrary{breakable,skins}
\usepackage{algorithm,algpseudocode}
\usepackage{placeins,xspace,dblfloatfix}

\usepackage{tikz}
\usetikzlibrary{arrows.meta,positioning,calc}
\definecolor{TableStripe}{HTML}{F2F2F2}
\definecolor{TheoremTint}{HTML}{FFF2F2}
\definecolor{TheoremRule}{HTML}{BF0000}
\definecolor{NoticeTint}{HTML}{FFF3DD}
\definecolor{FlowBlue}{HTML}{BF0000}
\definecolor{FlowGreen}{HTML}{606060}
\newtcolorbox{statementbox}{enhanced,colback=TheoremTint,frame hidden,borderline west={2.5pt}{0pt}{TheoremRule},boxrule=0pt,sharp corners,left=6pt,right=4pt,top=3pt,bottom=3pt}
\newtheorem{theorem}{Theorem}
\newtheorem{proposition}[theorem]{Proposition}

\newcommand{\E}{\mathbb{E}}
\newcommand{\Var}{\operatorname{Var}}
\newcommand{\Cov}{\operatorname{Cov}}

\newcommand{\TV}{\operatorname{TV}}
\newcommand{\ind}{\mathbb{I}}
\newcommand{\clip}{\operatorname{clip}}

\newcommand{\metric}[1]{\csname val@#1\endcsname}
\newcommand{\tabsetup}{\small\setlength{\tabcolsep}{3.5pt}\renewcommand{\arraystretch}{1.12}\rowcolors{2}{TableStripe}{white}}
\newcommand{\tablebundle}[1]{\begin{table*}[!tbp]
  \begingroup
  \renewenvironment{table}[1][]{}{}\renewenvironment{table*}[1][]{}{}#1
  \endgroup
  \end{table*}}
\expandafter\def\csname val@hp_pccc_jem\endcsname{56.71}
\expandafter\def\csname val@hp_pccc_f1\endcsname{78.69}
\expandafter\def\csname val@hp_pccc_em\endcsname{72.84}
\expandafter\def\csname val@hp_pccc_ucr\endcsname{6.16}
\expandafter\def\csname val@hp_am_jem\endcsname{52.52}
\expandafter\def\csname val@hp_am_f1\endcsname{78.22}
\expandafter\def\csname val@hp_am_em\endcsname{72.07}
\expandafter\def\csname val@hp_am_ucr\endcsname{8.40}
\expandafter\def\csname val@hp_cm_jem\endcsname{51.83}
\expandafter\def\csname val@hp_cm_f1\endcsname{76.94}
\expandafter\def\csname val@hp_cm_em\endcsname{70.51}
\expandafter\def\csname val@hp_cm_ucr\endcsname{9.31}
\expandafter\def\csname val@hp_c3_jem\endcsname{55.82}
\expandafter\def\csname val@hp_c3_f1\endcsname{79.43}
\expandafter\def\csname val@hp_c3_em\endcsname{73.62}
\expandafter\def\csname val@hp_c3_ucr\endcsname{11.07}
\expandafter\def\csname val@hp_gain_am\endcsname{4.19}
\expandafter\def\csname val@hp_gain_cm\endcsname{4.88}
\expandafter\def\csname val@hp_gain_c3\endcsname{0.89}
\expandafter\def\csname val@mu_pccc_jem\endcsname{36.82}
\expandafter\def\csname val@mu_pccc_f1\endcsname{62.82}
\expandafter\def\csname val@mu_pccc_em\endcsname{54.37}
\expandafter\def\csname val@mu_pccc_ucr\endcsname{6.28}
\expandafter\def\csname val@mu_am_jem\endcsname{32.96}
\expandafter\def\csname val@mu_am_f1\endcsname{60.40}
\expandafter\def\csname val@mu_am_em\endcsname{51.20}
\expandafter\def\csname val@mu_am_ucr\endcsname{8.95}
\expandafter\def\csname val@mu_cm_jem\endcsname{31.03}
\expandafter\def\csname val@mu_cm_f1\endcsname{59.12}
\expandafter\def\csname val@mu_cm_em\endcsname{49.62}
\expandafter\def\csname val@mu_cm_ucr\endcsname{9.59}
\expandafter\def\csname val@mu_c3_jem\endcsname{35.93}
\expandafter\def\csname val@mu_c3_f1\endcsname{63.06}
\expandafter\def\csname val@mu_c3_em\endcsname{54.39}
\expandafter\def\csname val@mu_c3_ucr\endcsname{11.25}
\expandafter\def\csname val@mu_gain_am\endcsname{3.86}
\expandafter\def\csname val@mu_gain_cm\endcsname{5.80}
\expandafter\def\csname val@mu_gain_c3\endcsname{0.89}
\expandafter\def\csname val@wiki_pccc_jem\endcsname{52.17}
\expandafter\def\csname val@wiki_pccc_f1\endcsname{76.64}
\expandafter\def\csname val@wiki_pccc_em\endcsname{70.52}
\expandafter\def\csname val@wiki_pccc_ucr\endcsname{6.37}
\expandafter\def\csname val@wiki_am_jem\endcsname{50.36}
\expandafter\def\csname val@wiki_am_f1\endcsname{76.28}
\expandafter\def\csname val@wiki_am_em\endcsname{69.96}
\expandafter\def\csname val@wiki_am_ucr\endcsname{8.46}
\expandafter\def\csname val@wiki_cm_jem\endcsname{50.36}
\expandafter\def\csname val@wiki_cm_f1\endcsname{76.17}
\expandafter\def\csname val@wiki_cm_em\endcsname{69.93}
\expandafter\def\csname val@wiki_cm_ucr\endcsname{9.82}
\expandafter\def\csname val@wiki_c3_jem\endcsname{52.38}
\expandafter\def\csname val@wiki_c3_f1\endcsname{76.96}
\expandafter\def\csname val@wiki_c3_em\endcsname{70.89}
\expandafter\def\csname val@wiki_c3_ucr\endcsname{11.34}
\expandafter\def\csname val@wiki_gain_am\endcsname{1.81}
\expandafter\def\csname val@wiki_gain_cm\endcsname{1.81}
\expandafter\def\csname val@wiki_gain_c3\endcsname{-0.21}
\expandafter\def\csname val@inf0_effect\endcsname{4.88}
\expandafter\def\csname val@inf0_ci\endcsname{[3.99, 5.76]}
\expandafter\def\csname val@inf0_p\endcsname{0.007812}
\expandafter\def\csname val@inf0_holm\endcsname{0.046875}
\expandafter\def\csname val@inf1_effect\endcsname{4.19}
\expandafter\def\csname val@inf1_ci\endcsname{[2.87, 5.58]}
\expandafter\def\csname val@inf1_p\endcsname{0.007812}
\expandafter\def\csname val@inf1_holm\endcsname{0.046875}
\expandafter\def\csname val@inf2_effect\endcsname{5.80}
\expandafter\def\csname val@inf2_ci\endcsname{[4.26, 7.41]}
\expandafter\def\csname val@inf2_p\endcsname{0.007812}
\expandafter\def\csname val@inf2_holm\endcsname{0.046875}
\expandafter\def\csname val@inf3_effect\endcsname{3.86}
\expandafter\def\csname val@inf3_ci\endcsname{[2.63, 5.13]}
\expandafter\def\csname val@inf3_p\endcsname{0.007812}
\expandafter\def\csname val@inf3_holm\endcsname{0.046875}
\expandafter\def\csname val@inf4_effect\endcsname{0.12}
\expandafter\def\csname val@inf4_ci\endcsname{[-0.17, 0.41]}
\expandafter\def\csname val@inf4_p\endcsname{0.328125}
\expandafter\def\csname val@inf4_holm\endcsname{0.328125}
\expandafter\def\csname val@inf5_effect\endcsname{0.36}
\expandafter\def\csname val@inf5_ci\endcsname{[-0.09, 0.79]}
\expandafter\def\csname val@inf5_p\endcsname{0.109375}
\expandafter\def\csname val@inf5_holm\endcsname{0.218750}
\expandafter\def\csname val@gain_correlation_cm\endcsname{0.120}
\expandafter\def\csname val@gain_correlation_am\endcsname{0.210}
\expandafter\def\csname val@factor_credit\endcsname{5.90}
\expandafter\def\csname val@factor_credit_ci\endcsname{[4.38, 7.44]}
\expandafter\def\csname val@factor_credit_ucr\endcsname{-3.28}
\expandafter\def\csname val@factor_cost\endcsname{0.78}
\expandafter\def\csname val@factor_cost_ci\endcsname{[0.24, 1.30]}
\expandafter\def\csname val@factor_cost_ucr\endcsname{-7.74}
\expandafter\def\csname val@factor_interaction\endcsname{-0.21}
\expandafter\def\csname val@factor_interaction_ci\endcsname{[-1.14, 0.65]}
\expandafter\def\csname val@factor_interaction_ucr\endcsname{-0.06}
\expandafter\def\csname val@ablation_search\endcsname{35.64}
\expandafter\def\csname val@loss_search\endcsname{1.18}
\expandafter\def\csname val@ablation_read\endcsname{36.14}
\expandafter\def\csname val@loss_read\endcsname{0.68}
\expandafter\def\csname val@ablation_verify\endcsname{35.73}
\expandafter\def\csname val@loss_verify\endcsname{1.10}
\expandafter\def\csname val@ablation_final\endcsname{36.13}
\expandafter\def\csname val@loss_final\endcsname{0.69}
\expandafter\def\csname val@ablation_uniform\endcsname{35.86}
\expandafter\def\csname val@loss_uniform\endcsname{0.97}
\expandafter\def\csname val@oracle_model_bias\endcsname{0.0353}
\expandafter\def\csname val@oracle_model_rmse\endcsname{0.1160}
\expandafter\def\csname val@oracle_model_nrmse\endcsname{0.586}
\expandafter\def\csname val@oracle_model_gradient_mse\endcsname{0.0984}
\expandafter\def\csname val@oracle_uncorrected_bias\endcsname{0.0266}
\expandafter\def\csname val@oracle_uncorrected_rmse\endcsname{0.0922}
\expandafter\def\csname val@oracle_uncorrected_nrmse\endcsname{0.465}
\expandafter\def\csname val@oracle_uncorrected_gradient_mse\endcsname{0.0564}
\expandafter\def\csname val@oracle_ips_bias\endcsname{-0.0027}
\expandafter\def\csname val@oracle_ips_rmse\endcsname{0.0690}
\expandafter\def\csname val@oracle_ips_nrmse\endcsname{0.348}
\expandafter\def\csname val@oracle_ips_gradient_mse\endcsname{0.0739}
\expandafter\def\csname val@oracle_uniform_bias\endcsname{0.0002}
\expandafter\def\csname val@oracle_uniform_rmse\endcsname{0.0649}
\expandafter\def\csname val@oracle_uniform_nrmse\endcsname{0.327}
\expandafter\def\csname val@oracle_uniform_gradient_mse\endcsname{0.0733}
\expandafter\def\csname val@oracle_adaptive_bias\endcsname{-0.0037}
\expandafter\def\csname val@oracle_adaptive_rmse\endcsname{0.0666}
\expandafter\def\csname val@oracle_adaptive_nrmse\endcsname{0.336}
\expandafter\def\csname val@oracle_adaptive_gradient_mse\endcsname{0.0682}
\expandafter\def\csname val@gradient_reduction\endcsname{7.04}
\expandafter\def\csname val@tail_adv_mean\endcsname{0.011}
\expandafter\def\csname val@tail_adv_var\endcsname{1.902}
\expandafter\def\csname val@tail_adv_p99\endcsname{6.965}
\expandafter\def\csname val@tail_adv_p999\endcsname{16.507}
\expandafter\def\csname val@tail_top1_mass\endcsname{25.927}
\expandafter\def\csname val@tail_predictor_mean\endcsname{0.045}
\expandafter\def\csname val@tail_predictor_var\endcsname{0.027}
\expandafter\def\csname val@tail_predictor_abs_p99\endcsname{0.446}
\expandafter\def\csname val@tail_mean_q\endcsname{0.137}
\expandafter\def\csname val@tail_cost_mape\endcsname{17.297}
\expandafter\def\csname val@tail_generation_ratio\endcsname{1.902}
\expandafter\def\csname val@tail_relative_gradient_variance\endcsname{1.000}
\expandafter\def\csname val@floor_low_jem\endcsname{35.73}
\expandafter\def\csname val@floor_high_jem\endcsname{36.91}
\expandafter\def\csname val@floor_low_var\endcsname{1.033}
\expandafter\def\csname val@floor_high_var\endcsname{0.921}
\expandafter\def\csname val@drift_default\endcsname{0.0099}
\expandafter\def\csname val@drift_default_ci\endcsname{[0.0092, 0.0106]}
\expandafter\def\csname val@hp_am_missing\endcsname{11.05}
\expandafter\def\csname val@hp_am_extra\endcsname{4.97}
\expandafter\def\csname val@hp_am_wrong\endcsname{3.54}
\expandafter\def\csname val@hp_am_correct_answer_wrong_support\endcsname{19.56}
\expandafter\def\csname val@hp_pccc_missing\endcsname{7.76}
\expandafter\def\csname val@hp_pccc_extra\endcsname{4.05}
\expandafter\def\csname val@hp_pccc_wrong\endcsname{4.32}
\expandafter\def\csname val@hp_pccc_correct_answer_wrong_support\endcsname{16.13}
\expandafter\def\csname val@mu_am_missing\endcsname{11.08}
\expandafter\def\csname val@mu_am_extra\endcsname{4.51}
\expandafter\def\csname val@mu_am_wrong\endcsname{2.64}
\expandafter\def\csname val@mu_am_correct_answer_wrong_support\endcsname{18.24}
\expandafter\def\csname val@mu_pccc_missing\endcsname{9.36}
\expandafter\def\csname val@mu_pccc_extra\endcsname{4.13}
\expandafter\def\csname val@mu_pccc_wrong\endcsname{4.06}
\expandafter\def\csname val@mu_pccc_correct_answer_wrong_support\endcsname{17.55}
\expandafter\def\csname val@lang_para_cm\endcsname{85.58}
\expandafter\def\csname val@lang_para_cm_count\endcsname{2054}
\expandafter\def\csname val@lang_para_am\endcsname{89.21}
\expandafter\def\csname val@lang_para_am_count\endcsname{2141}
\expandafter\def\csname val@lang_para_c3\endcsname{90.00}
\expandafter\def\csname val@lang_para_c3_count\endcsname{2160}
\expandafter\def\csname val@lang_para_pccc\endcsname{92.21}
\expandafter\def\csname val@lang_para_pccc_count\endcsname{2213}
\expandafter\def\csname val@lang_deletion_cm\endcsname{63.46}
\expandafter\def\csname val@lang_deletion_cm_count\endcsname{1523}
\expandafter\def\csname val@lang_deletion_am\endcsname{68.67}
\expandafter\def\csname val@lang_deletion_am_count\endcsname{1648}
\expandafter\def\csname val@lang_deletion_c3\endcsname{73.17}
\expandafter\def\csname val@lang_deletion_c3_count\endcsname{1756}
\expandafter\def\csname val@lang_deletion_pccc\endcsname{74.83}
\expandafter\def\csname val@lang_deletion_pccc_count\endcsname{1796}
\expandafter\def\csname val@lang_scope_cm\endcsname{50.58}
\expandafter\def\csname val@lang_scope_cm_count\endcsname{1214}
\expandafter\def\csname val@lang_scope_am\endcsname{55.88}
\expandafter\def\csname val@lang_scope_am_count\endcsname{1341}
\expandafter\def\csname val@lang_scope_c3\endcsname{62.83}
\expandafter\def\csname val@lang_scope_c3_count\endcsname{1508}
\expandafter\def\csname val@lang_scope_pccc\endcsname{62.25}
\expandafter\def\csname val@lang_scope_pccc_count\endcsname{1494}
\expandafter\def\csname val@lang_respond_am\endcsname{63.78}
\expandafter\def\csname val@lang_respond_pccc\endcsname{69.36}
\expandafter\def\csname val@citation_c3_str_em\endcsname{43.81}
\expandafter\def\csname val@citation_c3_precision\endcsname{81.19}
\expandafter\def\csname val@citation_c3_recall\endcsname{75.49}
\expandafter\def\csname val@citation_c3_tokens\endcsname{165.48}
\expandafter\def\csname val@citation_c3_citations\endcsname{5.44}
\expandafter\def\csname val@citation_c3_missed_facets\endcsname{2.22}
\expandafter\def\csname val@citation_pccc_str_em\endcsname{43.51}
\expandafter\def\csname val@citation_pccc_precision\endcsname{83.40}
\expandafter\def\csname val@citation_pccc_recall\endcsname{73.51}
\expandafter\def\csname val@citation_pccc_tokens\endcsname{150.91}
\expandafter\def\csname val@citation_pccc_citations\endcsname{5.39}
\expandafter\def\csname val@citation_pccc_missed_facets\endcsname{2.23}
\expandafter\def\csname val@human_am_rate\endcsname{9.17}
\expandafter\def\csname val@human_am_kappa\endcsname{0.654}
\expandafter\def\csname val@human_am_agreement\endcsname{93.89}
\expandafter\def\csname val@human_pccc_rate\endcsname{7.67}
\expandafter\def\csname val@human_pccc_kappa\endcsname{0.695}
\expandafter\def\csname val@human_pccc_agreement\endcsname{95.11}
\expandafter\def\csname val@noise20_am\endcsname{30.09}
\expandafter\def\csname val@noise20_pccc\endcsname{30.21}
\expandafter\def\csname val@compute_cm_flops_1e19\endcsname{1.339}
\expandafter\def\csname val@compute_cm_gpu_hours\endcsname{119.53}
\expandafter\def\csname val@compute_cm_wall_hours\endcsname{16.76}
\expandafter\def\csname val@compute_pccc_flops_1e19\endcsname{1.521}
\expandafter\def\csname val@compute_pccc_gpu_hours\endcsname{138.00}
\expandafter\def\csname val@compute_pccc_wall_hours\endcsname{21.10}
\expandafter\def\csname val@fixed_generation_ratio\endcsname{1.884}
\expandafter\def\csname val@fixed_flops_ratio\endcsname{2.030}
\expandafter\def\csname val@fixed_gpu_ratio\endcsname{2.227}

\title{Learning Which Evidence-Seeking Decisions Matter:\\Calibrated Counterfactual Replay for Language Agents}
\author{
\begin{tabular}{@{}cccc@{}}
Renwei Meng &
Chengyi Sun &
Hongyi Zhang &
Xiaoyi Hou \\[0.35em]
\normalfont\normalsize Anhui University &
\normalfont\normalsize \begin{tabular}[t]{@{}c@{}}University of Science and\\Technology Beijing\end{tabular} &
\normalfont\normalsize Anhui University &
\normalfont\normalsize Anhui University \\[0.35em]
\normalfont\scriptsize ORCID: \href{https://orcid.org/0009-0001-6879-6629}{0009-0001-6879-6629} &
\normalfont\scriptsize ORCID: \href{https://orcid.org/0009-0009-3869-0338}{0009-0009-3869-0338} &
\normalfont\scriptsize ORCID: \href{https://orcid.org/0009-0007-8809-381X}{0009-0007-8809-381X} &
\normalfont\scriptsize ORCID: \href{https://orcid.org/0009-0000-8271-6368}{0009-0000-8271-6368}
\end{tabular}
}
\hypersetup{pdftitle={},pdfauthor={},pdfsubject={},pdfkeywords={},pdfcreator={},pdfproducer={},pdftrapped={}}
\ifdefined\pdfinfoomitdate\pdfinfoomitdate=1\fi
\ifdefined\pdfsuppressptexinfo\pdfsuppressptexinfo=-1\fi
\ifdefined\pdftrailerid\pdftrailerid{}\fi

\begin{document}
\maketitle
\begin{abstract}
Language agents answer questions by searching, reading, and checking claims. A reward on the final answer says little about which of those decisions earned it, since an agent can reach the right string after redundant searches or with half the evidence missing. We treat the question as an experiment. At a saved decision point, PCCC-RL, for prefix-conditioned counterfactual credit, restores the tool state, runs the action the agent chose alongside a feasible alternative, and lets the same frozen policy finish both branches. The difference in final score then samples what the decision was worth. Replaying everything is too expensive, so the method predicts each action's credit up front, checks a random subset, and uses the recorded selection probabilities to correct the errors of that prediction, spending the budget where a further check can still move the training signal. Alternatives are drawn from the prefix alone, which keeps the comparison a usable gradient baseline. On HotpotQA and MuSiQue, joint exact match, requiring a correct answer together with complete supporting evidence, reaches \metric{hp_pccc_jem}\% and \metric{mu_pccc_jem}\%, gains of \metric{hp_gain_am} and \metric{mu_gain_am} points over a baseline with the same replay access. The gain concentrates in complete evidence chains and accumulates over many updates. Single-update effects stay uncertain, and citation coverage and scope distinctions such as ``none'' versus ``not all'' remain open.
\end{abstract}

\section{Introduction}
Consider a language agent asked, ``In which city was the founder of Cedar Museum born?'' One passage names Mara Lin as the founder. A second says that she was born in Bellhaven. Neither answers the question alone, and only a well-chosen sequence of moves will connect them: a search phrased to surface the second passage, a read that expands it, a verification that the text really does entail the birthplace claim. Retrieve the wrong passage first and the chain never closes. Recent work makes this sequencing part of the task rather than a preface to it, by scoring the supporting evidence and the citations along with the answer itself \citep{lewis2020rag,gao2023alce}.

A correct final answer is a poor teacher. It arrives once, at the end, after a dozen decisions of which some were essential, some redundant, and some actively misleading, and a single scalar cannot separate them. Scoring intermediate steps is the usual remedy, and it does provide more local feedback, but a step can look entirely reasonable and still change nothing about the outcome \citep{shao2024deepseekmath,huang2024citations}. What is missing is a notion of credit: for one decision, at one point in one episode, how much did it contribute to an answer that is both correct and fully supported?

We answer that question by experiment. Return to the decision point with the tool state restored, make a different feasible choice, and let the same frozen agent play out the rest of the episode. The gap between the two final scores samples what the decision was worth against the alternative actually offered. This makes credit a comparison rather than a verdict. Two searches that are equally useful come out equally credited, and no action is ever declared necessary in the abstract. It is also cheap in the currency that matters, because the alternative can be drawn from a pool of legal actions at the same prefix.

\begin{figure*}[!t]
\centering
\includegraphics[width=\textwidth]{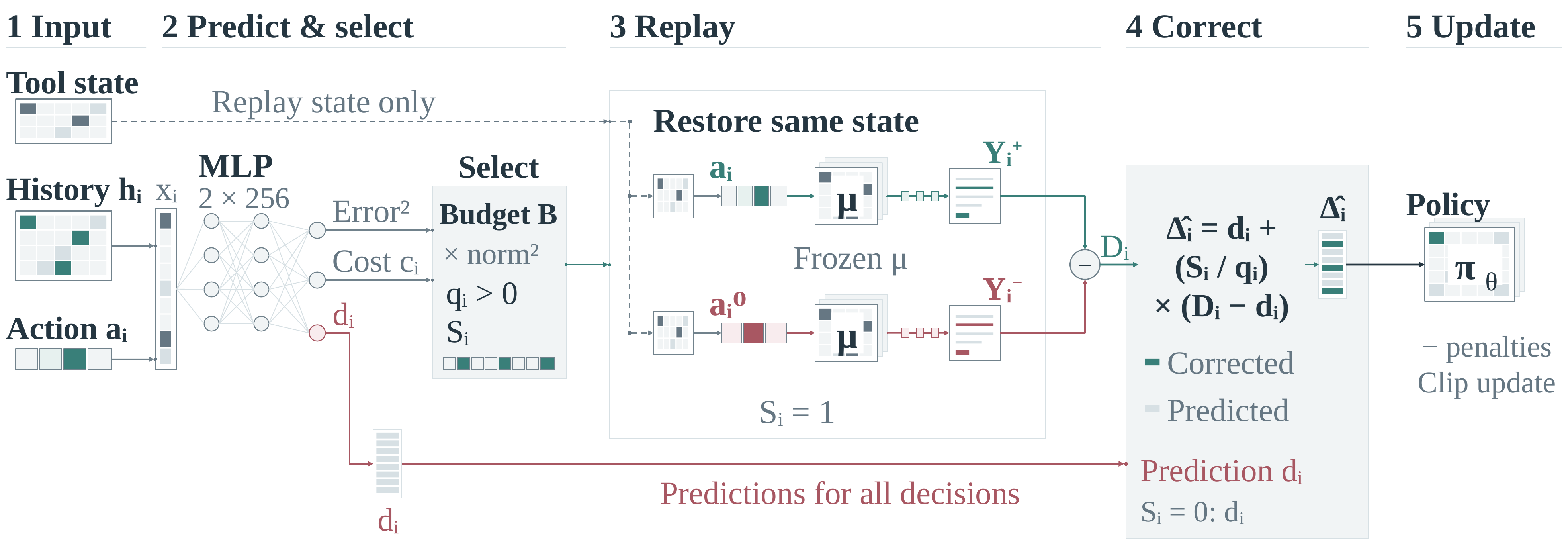}
\caption{One PCCC-RL iteration. A predictor fitted on earlier batches assigns every decision a provisional credit $d_i$, and predicted uncertainty, pair cost, and a score-norm proxy turn these into selection probabilities $q_i$ under budget $B$. Selected decisions replay from the restored tool state, once with the action taken and once with a prefix-only alternative, and the resulting score difference corrects the prediction before a single clipped, cost-adjusted update. Teal rows carry observed corrections and gray rows keep their predictions.}
\label{fig:flow}
\end{figure*}

Three things stand between this comparison and a training signal. The first is cost, since replaying every action doubles the work. The second is coverage, since replaying only some actions leaves the rest unobserved, and the actions we skip are not a random sample if we skipped them for a reason. The third is subtler and decides whether the signal teaches what we intend. If the alternative is chosen after inspecting the action it is meant to be compared against, the reference drifts with the action, and a decision that was genuinely bad can be made to look good. PCCC-RL, for prefix-conditioned counterfactual credit, handles all three at once: a predictor supplies credit for every action, randomized replay corrects the predictor wherever it happens to be checked, and the alternative is drawn from the prefix alone, using nothing that followed the decision. Replay effort follows predicted uncertainty, weighted by how much an update would care. Figure~\ref{fig:flow} shows one iteration end to end.

We then ask three separate questions of the resulting signal: is it accurate, does it improve behavior, and what does that accuracy cost? Accuracy is measured where the answer is known, in a finite decision environment whose futures can be enumerated. Behavior is measured on HotpotQA and MuSiQue, where the answer requires several linked facts. Cost is measured against every baseline we could match for information, in FLOPs and accelerator hours. Joint exact match rises by \metric{hp_gain_am} and \metric{mu_gain_am} points over a baseline given the same replay access, and the rise sits where the method aims, in complete evidence chains. We also say where the method does not win, in the sections where those numbers appear.

The contribution is a replay design that joins an executable comparison to a statistically corrected signal and a rule for spending the replay budget, together with the evidence that separates the three. The estimation and learning principles are classical. What is new is the assembly, the conditions under which it holds for recoverable agent interactions, and the experiments that keep accurate credit apart from useful updates. Proofs, settings, complete results, and statistics are in the appendices.

\section{Related Work}
\paragraph{Rewards on answers and on evidence.}
ReAct interleaves reasoning with tool use \citep{yao2023react}, and Search-R1 learns when to search from reinforcement feedback \citep{jin2025search}. Both score a completed response. Citation rewards give feedback at the level of individual claims \citep{huang2024citations}, and ALCE scores answer content together with citation quality \citep{gao2023alce}. Our question sits one level below all of these: not whether the final response was good, but whether this particular decision changed its fate.

\paragraph{Branching to compare decisions.}
Counterfactual baselines have a long history in multi-agent learning \citep{foerster2018coma}. C3 executes alternative continuations from a shared interaction trace \citep{chen2026c3}, TEMPO builds branching response trees to estimate intermediate values \citep{tran2026tempo}, and IBPO compares competing reasoning paths \citep{ding2026ibpo}. The design question we press on is what happens when only a fraction of branch outcomes is observed, which none of these methods has to answer.

\paragraph{Correcting what selective observation misses.}
Pairing predictions with inverse-probability corrections is a standard estimation strategy \citep{dudik2011dr,jiang2016dr,huang2020drpg}, and prefix-only baselines with constrained updates are equally standard \citep{schulman2017ppo,achiam2017cpo}. Bringing them to restored tool states raises issues that do not arise in a stateless bandit: the snapshot has to reproduce the environment exactly, the log must record selection probabilities faithfully, and the predictor has to be frozen before the labels it is judged on are collected. Theorems~\ref{thm:design} and~\ref{thm:gradient} state the identities that survive these constraints, and Propositions~\ref{prop:allocation} and~\ref{prop:approximation} characterize how the budget should be split when the weights are only estimated.

\section{Learning from Replayed Decisions}
\label{sec:method}
\subsection{Credit as a comparison}
The agent works with four actions. \textsc{Search} returns candidate passages for a query, \textsc{Read} expands a passage already observed, \textsc{Verify} asks whether some observed text entails a proposed claim, and \textsc{Final} closes the episode with an answer, a list of supporting identifiers, or an abstention. An episode runs from question to final response, and its history $h_t$ holds the question, every action taken, and everything the tools returned. The environment also carries hidden state $s_t$, such as internal tool counters and the identity of the retrieval index. We return to that distinction shortly, because what a restored snapshot restores is precisely the hidden part.

The final score $Y$ is one only when both the accepted answer and the complete supporting set are correct, so $Y=A_xE_x$ with $A_x$ and $E_x$ the two exact-match indicators. The agent learns a policy $\pi_\theta$ that maximizes $J(\theta)=\E_{\pi_\theta}[Y]$, subject to two expected-cost limits $J_C^j\le b_j$ on unsupported final claims and on attempted access to reserved evaluator fields. Appendix~\ref{app:assumptions} gives the formal process and the assumptions replay depends on.

To compare two decisions fairly, the agent that finishes both branches stays fixed within a training batch. At iteration $r$ that continuation policy is $\mu=\pi_{\theta_r}$. A proposal distribution $r_h$ chooses a feasible alternative from the question, the passages observed so far, and the legal tool schema. We call it \emph{prefix-only}, because it reads only that prefix: it never sees the chosen action, and it never sees the reward either branch went on to earn.

Let $Q_\mu(h,a)$ be the average final score after taking $a$ at history $h$ and continuing under $\mu$, and $B_\mu(h)$ the same average for an alternative drawn from $r_h$. Credit is the difference,
\begin{align}
 Q_\mu(h,a)&=\E[Y\mid h,\operatorname{do}(a),\mu],\nonumber\\
 B_\mu(h)&=\E_{a^0\sim r_h}[Q_\mu(h,a^0)],\label{eq:q}\\
 \Delta_\mu(h,a)&=Q_\mu(h,a)-B_\mu(h).\label{eq:credit}
\end{align}
Here $\operatorname{do}(a)$ means that the action is executed and not merely contemplated. Positive credit says the chosen action did better than the alternatives on offer; negative credit says it did worse. The number is relative to two moving parts, the alternative distribution and the frozen continuation, and we keep both in view whenever we interpret it.

\subsection{Correcting many predictions with a few replays}
A replay pair returns twice to the pre-action snapshot, executes $a_t$ and an independent $a_t^0\sim r_{h_t}$, and finishes both branches with $\mu$. The snapshot restores the tool state while the agent continues to see only its observable history, so the two branches begin from the same world and differ only in the action. We couple future random draws to reduce noise while preserving the correct marginal law for each branch, and we keep replay randomness independent of the draw that produced the original action. The observed difference is
\begin{equation}
 D_t=Y_t^+-Y_t^-,\qquad
 \E[D_t\mid h_t,a_t]=\Delta_\mu(h_t,a_t).
 \label{eq:pair}
\end{equation}
One pair is a noisy measurement, but its average recovers the credit of Equation~\ref{eq:credit}.

Two full continuations for every action would be prohibitive. A predictor trained on earlier batches instead gives each action a provisional credit $d_t=d_\phi(h_t,a_t,\mu)\in[-1,1]$, and the action is replayed with a known probability $q_t\in[q_{\min},1]$, the indicator $S_t$ recording the outcome with $S_t\sim\mathrm{Bernoulli}(q_t)$. Every action then receives
\begin{equation}
 \widehat\Delta_t=d_t+\frac{S_t}{q_t}(D_t-d_t).
 \label{eq:estimator}
\end{equation}
The rule starts from the prediction and, when a replay is drawn, adds the observed prediction error weighted by the inverse selection probability. Actions checked rarely receive large corrections when they are checked, while actions never checked cost nothing and keep their prediction. Averaging over the randomization recovers the target even though the predictor is imperfect, provided the predictor is frozen before the current batch's replay outcomes are collected, which is how we run it.

\begin{statementbox}
\begin{theorem}[Design expectation and variance]
\label{thm:design}
For fixed $(h,a)$, prediction $d$, and inclusion probability $q>0$, let $\Delta=\E[D\mid h,a]$ and $\sigma_D^2=\Var(D\mid h,a)$. Under the randomized replay design,
\begin{align}
 \E[\widehat\Delta\mid h,a]&=\Delta,\label{eq:unbiased}\\
 \Var(\widehat\Delta\mid h,a)
 &=\frac{\sigma_D^2}{q}+\left(\frac1q-1\right)(d-\Delta)^2.
 \label{eq:variance}
\end{align}
\end{theorem}
\end{statementbox}
The first line states that the correction removes systematic error in expectation. The second splits the remaining variance into two sources, the noise of the branch outcomes themselves and the error of the predictor, and the split matters because only the second responds to better prediction. A sharper predictor shrinks one term and leaves the other where it was, so replay never becomes free. When probabilities vary across a batch, the identity conditions on the complete allocation record available before replay begins. Appendix~\ref{app:proofs} gives the proof and those conditions.

\subsection{Why the reference has to be prefix-only}
Accurate comparisons do not by themselves produce the update we want. The alternative supplies the reference against which an action is rewarded, and if that reference moves with the sampled action, it can reverse the direction training takes. Restricting the reference to preceding history removes the dependence on the action, conditional on that history, and that single property is what makes the estimate usable as a gradient.

Write $\psi_t=\nabla_\theta\log\pi_\theta(a_t\mid h_t)|_{\theta_r}$ for the direction in which an action's log probability changes with the parameters. A reference that depends only on history has zero average contribution when multiplied by this vector, so subtracting it leaves the policy gradient of the original objective intact at the collection policy.
\begin{statementbox}
\begin{theorem}[Gradient-aligned credit]
\label{thm:gradient}
Under the conditions in Appendix~\ref{app:assumptions}, with $\mu=\pi_{\theta_r}$,
\begin{equation}
 \E\!\left[\sum_{t=0}^{H-1}\psi_t\widehat\Delta_t\right]
 =\nabla_\theta J(\theta_r).
 \label{eq:gradient}
\end{equation}
A reference $B(h,a)$ that reads the action adds the term $-\E[\sum_t\psi_t B(h_t,a_t)]$.
\end{theorem}
\end{statementbox}
This is an identity in expectation at the model that collected the batch. It is not a promise that any finite update improves anything, and Section~\ref{sec:results} reports a direct measurement of how far the two fall apart. It also does not license the evidence edits used later as diagnostics: those test how the agent responds to changed facts and are scored separately from the reference used in training.

Credit enters the update exactly once. Subtracting the two cost penalties gives the action's update signal, which we call its \emph{advantage},
\begin{equation}
 \widehat A_t=\widehat\Delta_t-
 \sum_{j=1}^{2}\lambda_j\{G_t^{C_j}-v_j(h_t)\}.
 \label{eq:advantage}
\end{equation}
Here $G_t^{C_j}$ is the remaining episode cost, $v_j$ predicts that cost from history, and $\lambda_j$ sets its price. The actor then takes a single clipped proximal-update epoch, which limits the incentive to move probabilities far in one step. The update uses complete-action probabilities under the same grammar mask, adjusts the cost multipliers, and refreshes $\mu$ for the next batch. Appendix~\ref{app:gradient-update} specifies the implemented objective.

\subsection{Spending the replay budget}
\label{sec:allocation}
A replay is worth most where prediction errors actually reach the update, their uncertainty is large, and the pair is cheap to run. For decision $i$ with pair cost $c_i>0$, the allocation weight combines learning sensitivity with the expected squared prediction residual,
\begin{equation}
 w_i=\|\psi_i\|^2\E[(D_i-d_i)^2\mid\mathcal F],
 \label{eq:weight}
\end{equation}
where $\mathcal F$ is the pre-replay allocation record. Minimizing the allocation-dependent gradient variance subject to an expected budget $\sum_i c_iq_i\le B$ gives
\begin{equation}
 q_i^*=\clip\!\left(\sqrt{\frac{w_i}{\eta c_i}},q_{\min},1\right).
 \label{eq:allocation}
\end{equation}
The multiplier $\eta$ is found by bisection so that the constraint binds. Weight pushes the probability up and cost pulls it down, and the positive floor $q_{\min}$ keeps every eligible decision observable at some rate. In practice the weights come from earlier batches and the score norm from a final-layer approximation, which is where estimation error enters. Propositions~\ref{prop:allocation} and~\ref{prop:approximation} derive the rule and bound the damage from misestimated weights.

Independent draws control the expected cost, so the realized cost can overrun the allowance. Selected jobs run to completion and we record the excess instead of cancelling it. Small probabilities also produce occasional very large corrections, and Section~\ref{sec:results} reports the tails, the clipping, and the failed updates instead of assuming that unbiased credit keeps training stable. Proposition~\ref{prop:concentration} and Appendix~\ref{app:drift} analyze the sampling noise and the slack introduced when the continuation policy moves between batches.

\begin{algorithm}[t]
\caption{One PCCC-RL training iteration}
\label{alg:training}
\begin{algorithmic}[1]
\State Freeze $\mu\gets\pi_{\theta_r}$ and collect agent interactions
\State Save pre-action tool states and actor token masks
\State Predict credit $d_i$, pair cost $c_i$, and weight $w_i$
\State Solve Equation~\ref{eq:allocation} for $q_i$ and draw $S_i$
\For{each selected decision $i$}
\State Draw the prefix-only alternative $a_i^0\sim r_{h_i}$
\State Restore the state twice and run both branches with $\mu$
\State Record score difference $D_i$ and execution cost
\EndFor
\State Correct credit and subtract costs using Equations~\ref{eq:estimator} and~\ref{eq:advantage}
\State Run one update epoch and adjust cost multipliers
\State Fit the next batch's predictors from the logged pairs
\end{algorithmic}
\end{algorithm}

\section{Experimental Setup}
\label{sec:experiments}
\paragraph{Tasks, evidence, and what counts as success.}
HotpotQA distractor, MuSiQue-Ans, and 2WikiMultiHopQA (2Wiki below) all require connected evidence from several passages \citep{yang2018hotpot,trivedi2022musique,ho2020two}. A deterministic local retrieval service exposes the context bundle released with each question, and the gold answers and support labels stay in a separate evaluator. Tuning uses training data and the official development sets are locked for evaluation. Answer F1 measures overlap with an accepted answer, while joint exact match (JEM) requires the answer and the complete evidence set, so an answer missing one supporting link fails. ALCE-ASQA supplies longer answers, the facts they are expected to cover, and citation support \citep{gao2023alce}.

\paragraph{What each comparison controls.}
All methods start from paired Qwen2.5-7B-Instruct models \citep{yang2025qwen} and share the interface and training settings, with the configuration given in Appendix~\ref{app:training}. Each reinforcement-learning run receives $10^8$ generated tokens, counted to include alternative proposals and replay continuations. \textbf{CM-RL} is the ordinary constrained baseline and spends the whole allowance on unreplayed trajectories. \textbf{AM-CF} is access-matched: it replays the same volume under the same schedule but fits a process score to the observed contrasts and uses that score as the advantage, with no inverse-probability correction at update time. \textbf{IM-CF} is information-matched: it receives the identical batch, replay records, selection indicators, probabilities, and lagged predictions for one update. SFT is the supervised warm start. The remaining rows of Table~\ref{tab:main} cover answer-based, search-based, and structured-credit baselines, each defined in Appendix~\ref{app:baselines}.

\paragraph{Separating estimation from behavior.}
EvidenceSwitch is a finite decision environment in which every future path can be enumerated, so exact average contributions are available without assuming that any predictor is correct. We use 2,000 saved decision points and 256 repeated replay designs per point, with a separate 4,096-pair reference checking 128 QA decision points. Behavior is tested with six evidence-edit families of 300 original--edited pairs each, evaluated under eight seeds, and a credit--constraint experiment that separates the training signal from the cost penalties. Saved checkpoints supply performance at intermediate budgets. Appendix~\ref{app:protocol} gives the full interface and evaluation protocols.

\paragraph{Uncertainty and paired tests.}
The primary experiments are repeated with eight paired seeds, matching initialization and data-order streams across methods. The six primary comparisons are the two main tasks against CM-RL and AM-CF, together with their identical-information single-update contrasts against IM-CF. Exact sign-flip tests assess the seed-level differences with Holm correction across this family at .05, and bootstrap intervals resample seeds and evidence-source clusters while keeping each method pair together \citep{dror2018significance,holm1979,owen2007}. Table~\ref{tab:inference} reports all six. A separate three-rater audit scores 1,800 outputs through 5,400 judgments. Appendix~\ref{app:statistics} gives the calculations.

\section{Results: From Credit to Supported Answers}
\label{sec:results}
\subsection{Credit tracks the exact contribution}
\begin{figure}[!tbp]
\centering\includegraphics[width=\columnwidth]{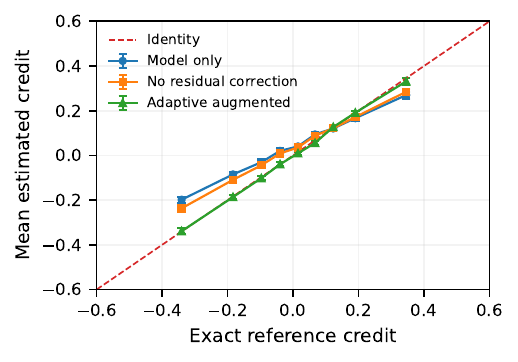}
\caption{Estimated credit against the exact contribution on 2,000 EvidenceSwitch decision points, averaged over 256 replay designs. The dashed line is perfect calibration, the bins hold equal numbers of points, and the bars are 1.96 standard errors of the binned mean.}
\label{fig:calibration}
\end{figure}
We measure accuracy first and separately from learning, because a credit signal that does not estimate what it claims would make every later result unreadable. EvidenceSwitch makes the exact contribution computable, so estimation error becomes something to measure instead of something to argue about. The predictor on its own shrinks toward zero and carries a visible offset (Figure~\ref{fig:calibration}), which is what a regression fit to a noisy target does, and its RMSE is \metric{oracle_model_rmse}. Adaptive replay correction brings the error down to \metric{oracle_adaptive_rmse} and straightens the curve (Table~\ref{tab:oracle}). Our reading is that the predictor's job is to set the scale of the estimate, not to rank individual decisions correctly, and that the ranking comes from the replays.

Where the budget goes matters more than how large it is. Uniform replay actually gives a slightly lower average credit RMSE, \metric{oracle_uniform_rmse}, while adaptive allocation cuts relative gradient squared error by \metric{gradient_reduction}\%. The two numbers measure different things, and the second is the one that matters: what steers an update is the score-weighted error, not the error averaged over decisions. A budget spent to reduce average prediction error is, in this setting, slightly misdirected, and we would rather say that plainly than report the metric that makes the allocation rule look better.

Individual corrections remain noisy, particularly when the selection probability is small. A decision replayed once in twenty can carry a very large correction when it is finally drawn, and the tails, the clipping behaviour, and the failed updates that follow are reported in Tables~\ref{tab:budget} and~\ref{tab:tails}, with the full estimator comparison in Appendix~\ref{app:results}.

\subsection{Do better credits produce better answers?}
\begin{table*}[!tbp]
\centering\tabsetup
\small
\begin{tabular}{lrrrrrr}
\toprule
Method & \multicolumn{2}{c}{HotpotQA} & \multicolumn{2}{c}{MuSiQue} & \multicolumn{2}{c}{2Wiki}\\
 & F1 & JEM & F1 & JEM & F1 & JEM\\
\midrule
SFT & 69.62 & $43.28\,\pm\,1.29$ & 49.18 & $22.93\,\pm\,2.37$ & 66.48 & $40.19\,\pm\,2.02$ \\
PPO-RLVR & 74.72 & $48.46\,\pm\,1.63$ & 55.69 & $28.57\,\pm\,1.70$ & 72.17 & $46.50\,\pm\,2.28$ \\
GRPO & 74.69 & $49.30\,\pm\,1.10$ & 56.71 & $29.50\,\pm\,1.54$ & 73.55 & $48.05\,\pm\,1.05$ \\
Search-R1$^\dagger$ & 77.21 & $52.06\,\pm\,1.10$ & 59.36 & $31.95\,\pm\,0.75$ & 74.97 & $48.21\,\pm\,2.07$ \\
TEMPO$^\dagger$ & 77.78 & $52.63\,\pm\,1.41$ & 60.99 & $33.25\,\pm\,1.82$ & 75.99 & $50.10\,\pm\,1.46$ \\
IBPO$^\dagger$ & 77.29 & $51.95\,\pm\,1.42$ & 61.41 & $33.84\,\pm\,2.43$ & 75.04 & $48.62\,\pm\,2.00$ \\
C3$^\dagger$ & 79.43 & $55.82\,\pm\,1.48$ & 63.06 & $35.93\,\pm\,2.34$ & 76.96 & $52.38\,\pm\,1.86$ \\
CM-RL & 76.94 & $51.83\,\pm\,1.00$ & 59.12 & $31.03\,\pm\,1.84$ & 76.17 & $50.36\,\pm\,2.09$ \\
AM-CF & 78.22 & $52.52\,\pm\,1.13$ & 60.40 & $32.96\,\pm\,1.64$ & 76.28 & $50.36\,\pm\,1.65$ \\
\textbf{PCCC-RL} & 78.69 & $56.71\,\pm\,1.05$ & 62.82 & $36.82\,\pm\,1.17$ & 76.64 & $52.17\,\pm\,0.83$ \\
\bottomrule
\end{tabular}
\caption{Grounded QA: answer F1 and JEM (\%). JEM is mean $\pm$ sample SD over eight paired seeds. All RL rows receive $10^8$ generated tokens, including reference proposals and both branch continuations. The SFT row reports the paired warm starts. $\dagger$ marks shared-interface adaptations. Each task retains its native support units.}
\label{tab:main}
\end{table*}

At the common token allowance, PCCC-RL reaches \metric{hp_pccc_jem}, \metric{mu_pccc_jem}, and \metric{wiki_pccc_jem} JEM on HotpotQA, MuSiQue, and 2Wiki (Table~\ref{tab:main}). Against CM-RL the gains on the first two tasks are \metric{hp_gain_cm} and \metric{mu_gain_cm} points, and against access-matched AM-CF they are \metric{hp_gain_am} and \metric{mu_gain_am}. The second comparison is the one we weight more heavily. AM-CF sees the same replay records under the same schedule and differs only in how it turns them into an advantage, so the remaining gap is attributable to the correction and not to extra information, which is exactly what the credit signal is supposed to provide.

The gain is not uniform, and the comparison with C3 is the place to be honest about it. C3 is ahead on answer F1 across all three tasks and ahead on 2Wiki JEM by a fifth of a point, while trailing ours on HotpotQA and MuSiQue by \metric{hp_gain_c3} and \metric{mu_gain_c3} points. We do not read the 2Wiki result as a tie. We read the pattern as an objective mismatch: C3 optimizes the return of a full trace, and answer strings are the part of the joint event that responds to that pressure, whereas our advantage over it sits entirely in whether the support set is complete. Removing the credit for search or for verification lowers MuSiQue JEM by \metric{loss_search} or \metric{loss_verify} points, while removing reading credit costs only \metric{loss_read} (Table~\ref{tab:ablations}). Reading is the action whose value is easiest to predict from the prefix, which is a plausible reason for the asymmetry and one we have not separated from action frequency.

\begin{table*}[!tbp]
\centering\tabsetup
\small
\begin{tabular}{lrrr}
\toprule
Task & IM-CF gain & Calibrated gain & Paired difference [95\% CI]\\
\midrule
HotpotQA & $0.45\,\pm\,0.31$ & $0.57\,\pm\,0.24$ & 0.12 [-0.17, 0.41] \\
MuSiQue & $0.18\,\pm\,0.45$ & $0.53\,\pm\,0.48$ & 0.36 [-0.09, 0.79] \\
\bottomrule
\end{tabular}
\caption{One update from an identical checkpoint and replay bank. Gains are mean $\pm$ SD in JEM points across eight seeds. Both methods see every branch label in the same bank. The last column compares their post-update predictions on the same evaluation items.}
\label{tab:bank-main}
\end{table*}

A stricter test freezes the checkpoint and hands the updater exactly the records IM-CF sees, so the two differ only in the arithmetic of the update. The advantage is \metric{inf4_effect} points on HotpotQA (95\% CI \metric{inf4_ci}) and \metric{inf5_effect} on MuSiQue (\metric{inf5_ci}), and neither interval excludes zero. This is the result we would most have wanted to be larger, and it is worth stating what it does and does not show: repeated online training moves JEM, while a single matched update leaves a gap too small to detect at eight seeds. Table~\ref{tab:bank} adds a detail we did not anticipate. Direct replay gains more in one update on MuSiQue, 0.82 points against 0.53, even though calibrated replay wins over the full run. Our explanation is variance: the raw contrast is unbiased but heavier-tailed, so it can win a single step by luck and lose the sequence, which is the same trade the allocation rule makes on purpose. Training trajectories and the complete single-update controls are in Appendix~\ref{app:results}.

\subsection{Answers follow the evidence, mostly}
\label{sec:language-results}
A benchmark score is worth more if the agent also responds correctly when the facts underneath it change. We test two behaviours apart: keeping the answer when an edit preserves meaning, and changing or declining when an edit removes what the evidence supports. These are evaluation edits only, and play no part in training.

\begin{figure*}[!tbp]
\centering
\includegraphics[width=\textwidth]{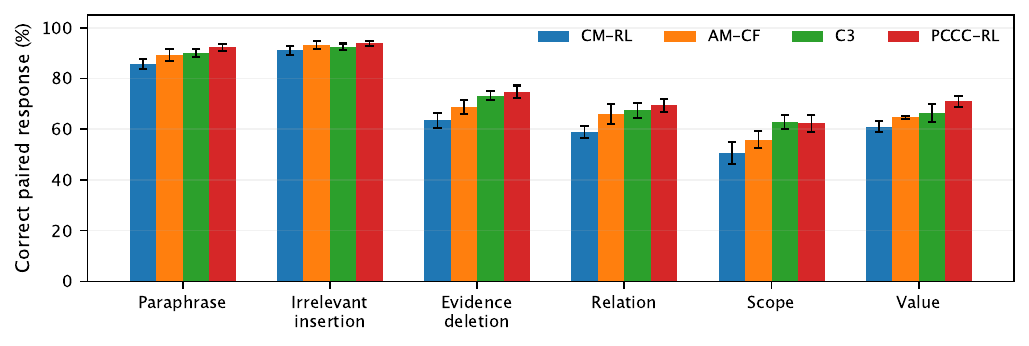}
\caption{Responses to preserved and changed evidence. Paraphrase and irrelevant insertion should leave the supported answer intact, while deletion, relation, scope, and value edits should change it. Each family has 300 unique pairs under eight seeds. Bars are means and error bars are between-seed SD.}
\label{fig:language}
\end{figure*}

\begin{figure*}[!tbp]
\centering
\includegraphics[width=.49\textwidth]{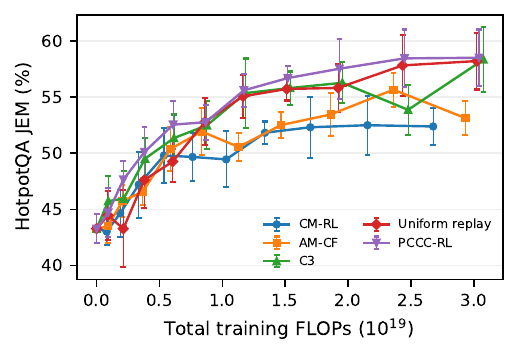}\hfill
\includegraphics[width=.49\textwidth]{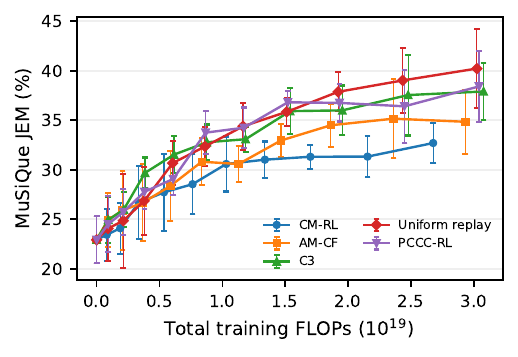}
\caption{Performance against total compute. Horizontal position is FLOPs including generation, input processing, updates, and auxiliary prediction. Points are eight-seed means with SD, at each method's own measured resource coordinates.}
\label{fig:frontier}
\end{figure*}
Paraphrase success reaches \metric{lang_para_pccc}\%. Across the four meaning-changing families, success averages \metric{lang_respond_pccc}\%, against \metric{lang_respond_am}\% for AM-CF (Figure~\ref{fig:language}). The deletion case is the clearest illustration. With the birthplace passage removed, the founder is still identified and the birthplace is no longer derivable, and retaining the old city would produce an answer the text does not support.

\begin{tcolorbox}[enhanced,colback=TableStripe,colbacktitle=TableStripe,coltitle=black,frame hidden,borderline west={2.5pt}{0pt}{FlowGreen},boxrule=0pt,sharp corners,title={When the birthplace evidence disappears},fonttitle=\small\bfseries,breakable,lines before break=8]
\small
\textbf{Evidence.} [A1] Mara Lin founded Cedar Museum. [A2] Mara Lin was born in Bellhaven.\\
\textbf{Edit.} Remove [A2], keeping [A1] and an unrelated museum-opening date.\\
\textbf{Required response.} The evidence no longer establishes the founder's birth city.\\
\textbf{AM-CF.} ``Bellhaven [A1].''\\
\textbf{PCCC-RL.} ``The supplied evidence identifies the founder but does not give her birthplace [A1].''
\end{tcolorbox}
Both systems answer Bellhaven before the edit. Afterwards, verification finds the missing bridge and PCCC-RL changes what it finalizes, which is what a correct response to a changed fact looks like. One case does not establish that every improved benchmark answer works this way, and we present it as an illustration of the mechanism rather than as evidence of its frequency.

Scope edits are the exception, and the exception is informative. PCCC-RL reaches \metric{lang_scope_pccc}\%, just below C3 at \metric{lang_scope_c3}\%, and Case C in Appendix~\ref{app:cases} shows why. Told that not all six sensors passed, the model concludes that some passed, which the sentence does not entail. Converting a negative universal into a positive existential is a quantifier inference that nothing in the terminal reward rewards directly. Better credit cannot supply a logical competence the data never exercises, and we would rather locate this failure in the model than attribute it to the estimator. Long answers show a different trade. On ALCE-ASQA, PCCC-RL attains \metric{citation_pccc_precision}\% citation precision, so attached citations more often support their claims, but \metric{citation_pccc_recall}\% recall against C3's \metric{citation_c3_recall}\%, so more claims go uncited; C3's precision is \metric{citation_c3_precision}\% and its longer outputs cover slightly more requested facts (Table~\ref{tab:citation}).

\subsection{What the gains cost}
Matching generated tokens does not match computation. At $10^8$ generated tokens, PCCC-RL uses \metric{compute_pccc_flops_1e19}$\times10^{19}$ floating-point operations against \metric{compute_cm_flops_1e19}$\times10^{19}$ for CM-RL, and \metric{compute_pccc_gpu_hours} accelerator-hours against \metric{compute_cm_gpu_hours} (Table~\ref{tab:compute}). The extra work goes to continuation, prediction, and restoring tool state. Held to the same factual training work, adaptive replay costs \metric{fixed_flops_ratio}$\times$ the FLOPs of terminal-reward training while consuming \metric{fixed_generation_ratio}$\times$ as many generated tokens (Table~\ref{tab:fixed-cost}).

The compute curves in Figure~\ref{fig:frontier} ask whether that work buys better supported answers, and the answer depends on where you stand on the curve. At the highest common MuSiQue compute point in Table~\ref{tab:resource-grid}, uniform replay and C3 both exceed PCCC-RL. Selective replay therefore buys an advantage in one regime and gives it back in another, which is a property of the allocation rule rather than a defect of the estimator, since more replay is not the same as better replay. Label quality is the sharper constraint: with 20\% of verifier labels corrupted, MuSiQue JEM falls to \metric{noise20_pccc}, near AM-CF's \metric{noise20_am} (Table~\ref{tab:noise}). No amount of statistical correction recovers from systematically wrong comparisons, and this is the failure mode we would expect to dominate in deployment.

\section{Conclusion}
Which intermediate decisions help an agent reach a correct answer with complete evidence is, in the end, an empirical question, and the contribution of this paper is a way to ask it. Restore the decision point, run two executable choices, and the difference between their outcomes is available for learning. A predictor with randomized correction supplies that difference wherever full replay is too expensive, and a prefix-only alternative keeps the resulting signal a valid policy gradient. On two multi-hop QA tasks the run produces better joint evidence correctness than any baseline given the same replay access, and the improvement appears in the part of the metric we designed it to move. The limits are equally specific: the gain takes many updates to appear, it does not extend to quantifier scope or to citation coverage, and it comes and goes with the compute budget. We think that is the right level of claim for a method whose assumptions can each be checked on its own, and the checks are the part we would most like to see repeated on other agent interfaces.
\label{sec:main-end}

\FloatBarrier
\section*{Limitations}
\phantomsection
\addcontentsline{toc}{section}{Limitations}
Replay rests on restoring tool state exactly. Live services and irreversible actions need isolation or a reproducible emulator, and the restoration audit in Appendix~\ref{app:oracle} covers only the fields we enumerate. The gradient identity holds in expectation at the collection policy, so it constrains nothing about an individual update, which is why we report the single-update contrast separately rather than folding it into the main claim. Inaccurate cost and residual predictions misallocate the budget, and small inclusion probabilities produce occasional corrections large enough to dominate a batch.

\begin{samepage}
The evidence tasks are predominantly English, and the annotated support sets may omit valid chains, which makes our JEM a lower bound on correct behaviour rather than an exact measure of it. Multilingual evidence seeking, and support discovery beyond the passages supplied with each question, need separate evaluation. Quantifier scope, coverage of multiple answer facts, and verifier errors all remain open. Compute comparisons depend on the model, the serving setup, and the budget, and the ordering we observe is not preserved across all three.
\par
\end{samepage}

\section*{Ethical Considerations}
\phantomsection
\addcontentsline{toc}{section}{Ethical Considerations}
Benchmark text retains its original licences and provenance. All tests run on disposable local services with isolated credentials. Human review uses de-identified outputs, an evidence-only rubric, and hidden method identity, and the logs keep actor-visible passages apart from protected evaluator labels. Consent, compensation, authorship, and assistance disclosures follow institutional and venue policy.

\phantomsection

\clearpage
\tableofcontents
\clearpage
\appendix
\section{Formal Assumptions and Complete Proofs}
\label{app:proofs}
The argument runs in three steps. Restored replay identifies the average difference between two choices; randomized correction recovers that difference while observing only some decisions; and a prefix-only reference then leaves the policy gradient where it was. The remaining propositions cover budget allocation, sampling error, and what happens as replay conditions degrade.
\subsection{State, randomness, and notation}
\label{app:assumptions}
The finite-horizon process is
$\mathcal M=(\mathcal S,\mathcal O,\mathcal A,p_0,P,O,H,Y,C)$.
The initial state is drawn from $p_0$, an action is followed by a transition under $P$ and an observation under $O$, and the actor receives all decision-relevant information through the observable history $h_t$. Actor randomization is independent of hidden simulator randomness conditional on that history. The terminal variable $Y$ is measurable with respect to the completed episode and takes values in $[0,1]$, and each episode cost is bounded in the same way. When an episode ends early, zero-score absorbing actions extend it to the fixed length $H$.

All expectations for an outer iteration condition on the training record preceding that iteration, which fixes the current parameters, continuation policy, proposal checkpoint, and lagged predictors. The state snapshot contains simulator state, tool state, retrieval-index identity, observation buffers, counters, and the relevant random-number-generator states; the actor sees none of the hidden fields. Restoring a snapshot and drawing a fresh future stream reproduces the continuation law conditional on the snapshot. Two branches may share future random numbers, but each must retain its correct marginal law, and future streams are independent of the draw that produced the factual action.

A reference action has distribution $r_h$ given the prefix. A finite pool comes from a fixed proposal checkpoint using only the prefix and the current legal schema, and $\nu(h,a^0)$ denotes prefix-observable schema and identifier validity. Under rejection sampling the accepted distribution is
\begin{equation}
 r_h(a^0)=\frac{\widetilde r_h(a^0)\nu(h,a^0)}
 {\sum_b\widetilde r_h(b)\nu(h,b)},
 \label{eq:reference-normalization}
\end{equation}
whose denominator must be positive. The finite-retry implementation appends a deterministic, always-valid abstention action once retries are exhausted, and that fallback is part of the declared distribution rather than an unrecorded dropped sample. Validity never depends on the terminal reward of the proposed branch. Pools are instantiated only for selected prefixes, from a separate random stream, under the same prefix-only rule.

For each candidate decision a potential paired observation $D_i$ is defined even when it is not executed, and the inclusion coin satisfies
\begin{equation}
 \begin{aligned}
 &\Pr(S_i=1\mid\mathcal F,D_1,\ldots,D_n)=q_i,\\
 &q_i\ge q_{\min}>0,
 \end{aligned}
 \label{eq:design-assumption}
\end{equation}
with $\mathcal F$ the complete allocation record available before replay. Inclusion coins are mutually independent conditional on this record. Costs, prefix features, and the predictor may all influence selection, but no selected job may be truncated without being logged, and the logged probabilities are the ones actually used to draw the coins rather than scores rounded or clipped afterwards.

The actor distribution is differentiable with a score of finite second moment, and its valid-action support is locally unchanged at $\theta_r$. Grammar masks are part of the sampling distribution. Derivatives and finite-horizon expectations interchange, and the predictor used in a batch is fixed before that batch's branch outcomes arrive. These are the standing assumptions for everything below.

\begin{table*}[t]
\centering\tabsetup
\begin{tabular}{p{.18\textwidth}p{.75\textwidth}}
\toprule
Symbol & Definition\\\midrule
$h_t,s_t,a_t$ & Observable history, hidden simulator state, and complete actor-generated action.\\
$\pi_\theta,\mu$ & Trainable sampling policy and frozen rollout policy $\mu=\pi_{\theta_r}$.\\
$r_h,a_t^0$ & Prefix-only reference distribution and a sampled reference action.\\
$Y,C^j,b_j$ & Terminal success, bounded episode cost, and the corresponding expected-cost limit.\\
$Q_\mu,B_\mu,\Delta_\mu$ & Factual continuation value, prefix-only reference value, and their difference.\\
$Y_t^+,Y_t^-,D_t$ & Paired branch returns and the difference $Y_t^+-Y_t^-$.\\
$S_t,q_t,d_t$ & Replay inclusion indicator, logged inclusion probability, and lagged credit prediction.\\
$\psi_t,\widehat A_t$ & Score vector of the complete sampled action and constrained credit advantage.\\
$c_i,w_i,B,\eta$ & Predicted pair cost, gradient-weighted residual second moment, expected replay budget, and allocation multiplier.\\
$\mathcal F$ & Pre-replay allocation record, including the selected checkpoint identities and sampling design.\\
\bottomrule
\end{tabular}
\caption{Notation. The proposal distribution $r_h$ and the outer-iteration index $r$ have distinct roles, and the subscript $h$ always identifies the former.}
\label{tab:notation}
\end{table*}

\subsection{Why replay estimates the intended difference}
Fix a history $h$ and a factual action $a$, and let $p(s\mid h)$ be the hidden-state distribution induced by collection with $\mu$. Conditional independence of actor and hidden-state randomization gives $p(s\mid h,a)=p(s\mid h)$. Write $V_\mu(s,h,a)$ for the expected terminal success after applying $a$ to snapshot state $s$ and continuing with $\mu$. Faithful replay then implies
\begin{align}
 \E[Y^+\mid h,a]
  &=\int V_\mu(s,h,a)p(ds\mid h),\nonumber\\
 \E[Y^-\mid h,a]
  &=\int\sum_b r_h(b)V_\mu(s,h,b)p(ds\mid h).
 \label{eq:identify-integrals}
\end{align}
The first integral is $Q_\mu(h,a)$ and the second is $B_\mu(h)$, so subtracting proves Equation~\ref{eq:pair}. Hidden state need not be reconstructed by a text-only model, because the replay service preserves it and the average runs under its collection distribution. Fixing a single captured snapshot instead conditions the contrast on that snapshot, and averaging over snapshots at a given history recovers the history-conditional quantity.

Shared random numbers change the joint law of $(Y^+,Y^-)$ without touching either expectation, and consequently
\begin{equation}
 \begin{aligned}
 \Var(D\mid h,a)&=\Var(Y^+\mid h,a)\\
 &\quad+\Var(Y^-\mid h,a)\\
 &\quad-2\Cov(Y^+,Y^-\mid h,a).
 \end{aligned}
 \label{eq:coupling-variance}
\end{equation}
Positive covariance reduces variance and negative covariance increases it, and we measure this covariance on development prefixes when choosing the coupling scheme.

\subsection{Proof of Theorem~\ref{thm:design}}
Fix $(h,a)$ and a fixed $q$, and let $e=D-d$. By design $\E[S\mid D,h,a]=q$ and $S^2=S$, so
\begin{align}
 \E[d+Se/q\mid h,a]
 &=d+\E[e\mid h,a]\nonumber\\
 &=\E[D\mid h,a]=\Delta.
\end{align}
For the variance, $d$ is fixed under the conditioning, hence
\begin{align}
 \Var(\widehat\Delta\mid h,a)
 &=\E[S e^2/q^2\mid h,a]\nonumber\\
 &\quad-\{\E[Se/q\mid h,a]\}^2\nonumber\\
 &=q^{-1}\E[e^2\mid h,a]-(\Delta-d)^2\nonumber\\
 &=q^{-1}\sigma_D^2+(q^{-1}-1)(\Delta-d)^2,
\end{align}
which proves both claims. For a batch-adaptive $q_i=q_i(\mathcal F)$, condition first on $\mathcal F$ and all potential $D_i$; Equation~\ref{eq:design-assumption} gives $\E[\widehat\Delta_i\mid\mathcal F,D_i]=D_i$, and a second expectation conditional on $(h_i,a_i)$ yields the same unbiasedness. The total variance in that case includes the additional mixing over $\mathcal F$, and the displayed pointwise formula is not substituted without the corresponding conditioning.

The model-only estimate carries conditional bias $d-\Delta$, while the inverse-inclusion estimate $SD/q$ is the special case $d=0$ with variance $\sigma_D^2/q+(q^{-1}-1)\Delta^2$. Augmentation therefore beats that special case exactly when $(d-\Delta)^2<\Delta^2$ for $q<1$, which means it is not an unconditional variance reduction for an arbitrarily poor predictor. This is the reason the predictor has to be reasonable before the correction helps, and the reason we measure its calibration instead of assuming it.

\paragraph{Altering logged probabilities.}
Suppose the actual inclusion probability is $q$ but the residual denominator is changed to $\bar q$. Then
\begin{equation}
 \E\!\left[d+\frac S{\bar q}(D-d)\right]-\Delta
   =\left(1-\frac q{\bar q}\right)(d-\Delta),
 \label{eq:clipping-bias}
\end{equation}
so any mismatch between the probability we report and the probability we used becomes bias proportional to the prediction error. The implementation bounds $q$ inside the sampling design and uses that exact probability in the denominator, and the all-step prediction stays available at both selected and unselected decisions.

\paragraph{Fitting the predictor.}
For a new predictor $d_\phi$ the lagged training loss is
\begin{equation}
 \widehat L(\phi)=\frac1n\sum_i\frac{S_i}{q_i}
          \{D_i-d_\phi(h_i,a_i,\mu_i)\}^2,
 \label{eq:critic-loss}
\end{equation}
which targets population squared error across eligible decisions rather than the distribution over heavily sampled ones. The residual second-moment predictor uses the same inclusion weights together with older out-of-fold predictions. Policy-checkpoint identity is part of the input, stale entries receive the recency weighting given in Appendix~\ref{app:training}, and fitting on a batch's own labels happens only after that batch's actor update.

\subsection{Proof of Theorem~\ref{thm:gradient}}
The likelihood of a trajectory factors into an initial-state term, action probabilities, and environment terms, and only the action probabilities depend on $\theta$. The likelihood-ratio identity therefore gives
\begin{equation}
 \begin{aligned}
 \nabla J(\theta_r)&=\E_{\mu}\left[Y\sum_{t=0}^{H-1}\psi_t\right]\\
 &=\E_\mu\left[\sum_t\psi_t Q_\mu(h_t,a_t)\right].
 \end{aligned}
 \label{eq:score-proof}
\end{equation}
For every fixed history, differentiability and normalization imply
\begin{equation}
 \begin{aligned}
 \E[\psi_t\mid h_t]
 &=\sum_a\mu(a\mid h_t)\nabla\log\pi_\theta(a\mid h_t)|_{\theta_r}\\
 &=\nabla\sum_a\pi_\theta(a\mid h_t)|_{\theta_r}=0.
 \end{aligned}
\end{equation}
Because $B_\mu(h_t)$ depends on history alone, its contribution to the score expectation vanishes, and subtracting it from Equation~\ref{eq:score-proof} leaves $\E[\sum_t\psi_t\Delta_\mu(h_t,a_t)]$. Theorem~\ref{thm:design} with iterated expectation then permits replacing $\Delta_\mu$ by $\widehat\Delta_t$, which proves Equation~\ref{eq:gradient}.

If the reference is $B(h_t,a_t)$ the corresponding term cannot be factored out of the conditional action expectation, and direct subtraction leaves
\begin{equation}
 \begin{aligned}
 &\E\left[\sum_t\psi_t\{Q_\mu-B\}\right]\\
 &\quad=\nabla J(\theta_r)\\
 &\qquad-\E\left[\sum_t\psi_tB(h_t,a_t)\right].
 \end{aligned}
 \label{eq:action-dependent-error}
\end{equation}
A minimal counterexample makes the failure concrete. Take one state, two actions, $Y=a\in\{0,1\}$, and a Bernoulli-logit actor with success probability $p$. The true derivative is $p(1-p)$. If the reference transformation returns the sampled action itself, so that $B(a)=a$, the contrast is identically zero and the substituted update vanishes. An accurate intervention contrast can therefore still be the wrong gradient signal, which is why the prefix-only restriction is a requirement and not a convenience.

For costs the same likelihood-ratio argument gives $\nabla J_C^j=\E[\sum_t\psi_t G_t^{C_j}]$, and a prefix-only $v_j$ again has zero score expectation. Combining these identities with fixed $\lambda_j$ shows that Equation~\ref{eq:advantage} targets the gradient of $J-\sum_j\lambda_jJ_C^j$ at collection, the constants $\lambda_jb_j$ contributing nothing to the actor derivative.

\paragraph{Implemented proximal update.}
\label{app:gradient-update}
The actor takes a clipped proximal update using the corrected advantage of Equation~\ref{eq:advantage}, with objective
\begin{equation}
 \E_{\tau\sim\mu}\!\left[\sum_t
 \min\{R_t\widehat A_t,\overline R_t\widehat A_t\}\right],
 \label{eq:ppo}
\end{equation}
where $R_t=\pi_\theta(a_t\mid h_t)/\mu(a_t\mid h_t)$ and $\overline R_t=\clip(R_t,1-\epsilon,1+\epsilon)$. The ratio compares new and old probabilities for the complete action, so clipping limits the incentive to move far in one step. Log probabilities include the actor-generated tokens under the same grammar mask. Cost multipliers follow projected ascent on $J_C^j-b_j$, and after one update epoch we refresh $\mu$ for the next batch.

\paragraph{Why credit enters once.}
An additional reward term $\alpha\Delta_t$ changes the return being optimized, whereas Equation~\ref{eq:gradient} already uses the credit as a replacement gradient signal for the original terminal objective; adding the term while also inserting $\Delta_t$ into the advantage defines a different update. That alternative is what the corresponding ablation measures against the single-credit default.

\subsection{Allocating an expected replay budget}
\label{app:allocation}
\begin{statementbox}
\begin{proposition}[Variance--cost allocation]
\label{prop:allocation}
Condition on a fixed allocation record $\mathcal F$, score vectors $\psi_i$, and predictions $d_i$. Let $w_i=\|\psi_i\|^2\E[(D_i-d_i)^2\mid\mathcal F]$ and $c_i>0$. Suppose $q_i\in[q_{\min},1]$ and $\sum_i c_iq_i\le B$. If $q_{\min}\sum_i c_i\le B$, the allocation-dependent trace variance is minimized by Equation~\ref{eq:allocation} for positive weights. Zero-weight entries can be assigned $q_{\min}$, and the multiplier is chosen to exhaust the budget unless all positive-weight entries already have $q_i=1$.
\end{proposition}
\end{statementbox}
\begin{proof}
Write the gradient estimate as $\widehat g=\sum_i\psi_i[d_i+S_i(D_i-d_i)/q_i]$. Conditional independence of the inclusion coins makes the cross-decision covariance terms independent of the $q_i$: for $i\ne j$ the factor $\E[S_iS_j\mid\mathcal F,D]/(q_iq_j)$ is one. Individual trace-variance terms are $w_i/q_i-\|\psi_i\|^2\{\E[D_i-d_i\mid\mathcal F]\}^2$, and the second term is again independent of the allocation, so it suffices to solve
\begin{equation}
 \begin{aligned}
 &\min_{q}\sum_i\frac{w_i}{q_i},\\
 &q_{\min}\le q_i\le1,\qquad\sum_i c_iq_i\le B.
 \end{aligned}
 \label{eq:convex-allocation}
\end{equation}
For $w_i>0$ the objective is strictly convex in $q_i$, since its second derivative $2w_i/q_i^3$ is positive. An interior Lagrangian stationary point satisfies $-w_i/q_i^2+\eta c_i=0$, giving $q_i=\sqrt{w_i/(\eta c_i)}$, and the box constraints project this onto $[q_{\min},1]$. Convexity makes the Karush--Kuhn--Tucker conditions sufficient. The cost is continuous and nonincreasing in $\eta$, so bisection reaches the active-budget solution. Zero weights cannot improve the objective by raising $q_i$, and the floor suffices. If $B<q_{\min}\sum_i c_i$ no feasible allocation exists and sampling must wait for a larger budget or a redesigned floor.
\end{proof}

The weights in this proposition condition on the complete allocation record, whereas Equation~\ref{eq:weight} is the prefix-local approximation actually predicted. Predictor error costs efficiency but not the expectation result, because the realised $q_i$ is logged and used. For an approximation $\widetilde w_i$ the finite-design excess objective is exactly
\begin{equation}
 \sum_i w_i\left(\frac1{\widetilde q_i}-\frac1{q_i^*}\right)\ge0,
\end{equation}
which the exact-reference environment evaluates directly.

\subsection{Using estimated allocation weights}
\begin{statementbox}
\begin{proposition}[Multiplicative allocation error]
\label{prop:approximation}
Let $\mathcal Q$ be the feasible set in Equation~\ref{eq:convex-allocation}, with the same costs, allowance, and floor for both designs, and define $F_w(q)=\sum_iw_i/q_i$. Suppose $0\le\epsilon<1$ and $(1-\epsilon)w_i\le\widetilde w_i\le(1+\epsilon)w_i$ for every entry, consistent zero weights included. If $q^*$ minimizes $F_w$ and $\widetilde q$ minimizes $F_{\widetilde w}$ on $\mathcal Q$, then
\begin{equation}
 F_w(\widetilde q)\le\frac{1+\epsilon}{1-\epsilon}F_w(q^*).
 \label{eq:approximation-bound}
\end{equation}
\end{proposition}
\end{statementbox}
\begin{proof}
Every $q_i$ is positive, so the coordinatewise weight bounds give $(1-\epsilon)F_w(q)\le F_{\widetilde w}(q)\le(1+\epsilon)F_w(q)$ at every feasible $q$. Optimality of $\widetilde q$ under the approximate weights then yields
\begin{align*}
 (1-\epsilon)F_w(\widetilde q)
 &\le F_{\widetilde w}(\widetilde q)\\
 &\le F_{\widetilde w}(q^*)\\
 &\le(1+\epsilon)F_w(q^*),
\end{align*}
and dividing by $1-\epsilon>0$ proves the result. This objective is exactly the allocation-dependent term isolated in Proposition~\ref{prop:allocation}, the remaining covariance and mean terms being fixed across feasible allocations.
\end{proof}
The bound translates score-proxy and residual-prediction error into a multiplicative loss under a shared cost constraint. Predicted-cost error enters differently, through realised budget utilization in Table~\ref{tab:budget}, and separating the two keeps allocation quality and cost fidelity independently inspectable.

\subsection{Bounding the error from selective replay}
\begin{statementbox}
\begin{proposition}[Projected replay-design tail bound]
\label{prop:concentration}
Condition on $\mathcal F$, all potential contrasts $D_i$, predictions $d_i$, and score vectors $\psi_i$. Let $e_i=D_i-d_i$ and let the inclusion coins be independent Bernoulli variables with probabilities $q_i>0$. Write $\widehat g=\sum_i\psi_i\widehat\Delta_i$ and $g_{\rm full}=\sum_i\psi_iD_i$. For any fixed unit vector $u$, define
\begin{align}
 V_u&=\sum_i(u^\top\psi_i)^2e_i^2(q_i^{-1}-1),\\
 M_u&=\max_i\frac{|u^\top\psi_i e_i|}{q_i}.
\end{align}
Then for $0<\alpha<1$, with conditional probability at least $1-\alpha$,
\begin{equation}
 \begin{split}
 |u^\top(\widehat g-g_{\rm full})|
 \le{}&\sqrt{2V_u\log(2/\alpha)}\\
 &+\tfrac23M_u\log(2/\alpha).
 \end{split}
 \label{eq:concentration}
\end{equation}
For an average gradient both sides are divided by the number of decisions. If $D_i,d_i\in[-1,1]$, then $M_u\le2\max_i\|\psi_i\|/q_{\min}$.
\end{proposition}
\end{statementbox}
\begin{proof}
Set $Z_i=(u^\top\psi_i)e_i(S_i/q_i-1)$. Then $\sum_iZ_i=u^\top(\widehat g-g_{\rm full})$, $\E[Z_i]=0$, $|Z_i|\le M_u$, and
\[
 \sum_i\E[Z_i^2]=\sum_i(u^\top\psi_i)^2e_i^2(q_i^{-1}-1)=V_u.
\]
Boundedness gives $|\E[Z_i^k]|\le M_u^{k-2}\E[Z_i^2]$ for $k\ge2$. Expanding the moment-generating function with $k!\ge2\,3^{k-2}$ then yields, for $0<\lambda<3/M_u$,
\begin{align*}
 \E[e^{\lambda Z_i}]
 &\le 1+\frac{\lambda^2\E[Z_i^2]}{2}
       \sum_{j=0}^\infty(\lambda M_u/3)^j,\\
 \log\E[e^{\lambda\sum_iZ_i}]
 &\le\frac{\lambda^2V_u}{2(1-\lambda M_u/3)},
\end{align*}
the second line following from independence by summing log moment-generating functions and using $\log(1+x)\le x$. Chernoff's inequality with $\lambda=t/(V_u+M_ut/3)$ gives
\[
 \Pr\{\sum_iZ_i\ge t\}\le
 \exp\!\left\{-\frac{t^2}{2(V_u+M_ut/3)}\right\}.
\]
Applying the same argument to $-Z_i$ and setting $x=\log(2/\alpha)$ with $t=\sqrt{2V_ux}+2M_ux/3$, direct expansion gives $t^2\ge2x(V_u+M_ut/3)$, so the two tail probabilities sum to at most $2e^{-x}=\alpha$. The zero-variance case is deterministic. Finally $|e_i|\le2$ and Cauchy--Schwarz give the displayed bound on $M_u$.
\end{proof}
The variance term measures how much residual uncertainty sits on decisions assigned small probabilities, and the maximum term measures sensitivity to the single largest weighted residual. That decomposition is why we report both average gradient variance and extreme advantage quantiles at a fixed allowance rather than one summary.

\subsection{Policy drift and restored state}
\label{app:drift}
\begin{statementbox}
\begin{proposition}[Frozen-policy drift]
\label{prop:drift}
Assume a terminal score in $[0,1]$ and at most $L$ remaining continuation decisions after the intervened action. If $\sup_h\TV(\pi(\cdot\mid h),\mu(\cdot\mid h))\le\varepsilon$ on every history reachable by either branch, then
\begin{align}
 |Q_\pi(h,a)-Q_\mu(h,a)|
 &\le1-(1-\varepsilon)^L\le L\varepsilon,\nonumber\\
 |\Delta_\pi(h,a)-\Delta_\mu(h,a)|
 &\le2\min\{1,L\varepsilon\}.
 \label{eq:drift-bound}
\end{align}
The reference distribution is held fixed in both contrasts.
\end{proposition}
\end{statementbox}
\begin{proof}
Couple actions from $\pi$ and $\mu$ maximally whenever their histories agree, and use identical environment randomness until an action differs. At each of the at most $L$ decisions the conditional agreement probability is at least $1-\varepsilon$, so the coupled paths differ with probability at most $1-(1-\varepsilon)^L$. Identical paths have identical terminal scores, while different paths change a bounded score by at most one, which proves the first inequality. The union bound gives $1-(1-\varepsilon)^L\le L\varepsilon$. Apply the same argument to the reference value and use the triangle inequality for the contrast, with the trivial range bound of two.
\end{proof}

When a uniform per-history KL bound $\kappa$ is available, Pinsker's inequality gives $\varepsilon\le\sqrt{\kappa/2}$. The implementation reports credit drift across updates and refreshes $\mu$ every outer iteration, and fresh branch sampling carries a continuation likelihood ratio of one.

\begin{proposition}[Replay-fidelity contribution]
\label{prop:fidelity}
Suppose the full continuation-path law under replay differs from the intended continuation law by at most $\delta$ in total variation, separately for the factual and reference branches. Then the replay contrast differs from the intended contrast by at most $2\delta$, and, combined with the preceding drift condition, the discrepancy from the current-policy contrast is at most $2\delta+2\min\{1,L\varepsilon\}$.
\end{proposition}
\begin{proof}
For any $Y\in[0,1]$ the difference of expectations under two probability laws is bounded by their total variation. Apply this once to each branch, add, and use a triangle inequality with Proposition~\ref{prop:drift}.
\end{proof}
The restoration audit checks the state fields and next-observation equality specified in Appendix~\ref{app:oracle}, with a binomial interval for that diagnostic.

\subsection{Uncertainty in a sampled intervention reference}
For independent branch pairs $D_1,\ldots,D_M\in[-1,1]$ with mean $\bar D_M$, Hoeffding's inequality gives
\begin{equation}
 \Pr\{|\bar D_M-\Delta|>u\}\le2\exp(-Mu^2/2).
\end{equation}
A pointwise $(1-\alpha)$ interval has half-width $\sqrt{2\log(2/\alpha)/M}$, clipped to the parameter range, and $N$ reference prefixes replace $\alpha$ by $\alpha/N$ for simultaneous control by a union bound. Pairing within each observation is allowed, whereas independent future streams are required across observations. We report this finite-sample uncertainty for the neural-agent Monte Carlo references; EvidenceSwitch instead uses exact finite-state dynamic programming and carries no Monte Carlo error in its oracle values.

\section{Experimental Interfaces and Reproducibility}
\label{app:protocol}
This appendix defines what the agent can observe and do, how the methods share training resources, and how each diagnostic is scored. The main text introduces why these comparisons exist; here we specify the settings needed to reproduce them.
\subsection{Tasks, splits, and terminal evaluation}
The three primary QA tasks use the context bundles distributed with HotpotQA distractor, MuSiQue-Ans, and 2WikiMultiHopQA. The evaluation sets contain 7,405, 2,417, and 12,576 questions \citep{yang2018hotpot,trivedi2022musique,ho2020two}, and tuning uses 3,000 held-out training questions per task drawn from a pool of up to 48,000. Questions that share a source-document component are assigned to the same partition when the tuning set is formed, and the development set is sealed before method selection. A task-specific actor is trained for each dataset. Equal seed identifiers pair initialization and data-order streams, not learned parameters across tasks.

The local interface exposes four complete actions. \textsc{Search} accepts a text query and returns up to five passages from the current question's context bundle. \textsc{Read} expands an observed document or sentence identifier. \textsc{Verify} requests entailment of a proposed claim from observed text. \textsc{Final} returns an answer, a list of supporting identifiers, and an abstention flag. Search runs against a fixed lexical index with deterministic tie breaking, which isolates the choice and interpretation of evidence inside the supplied bundle. The gold support set, answer aliases, and scoring rules live in a separate evaluator process.

\paragraph{Answer and support scoring.}
Answer EM uses the benchmark's normalization and accepted aliases, and token F1 uses the corresponding official normalization. HotpotQA and 2Wiki support is scored at the annotated supporting-fact granularity, while MuSiQue uses the supplied supporting-paragraph identities; predicted identifiers are canonicalized and deduplicated before the sets are compared. With $\mathcal S_x$ the accepted support sets for question $x$,
\begin{equation}
 E_x=\ind\{\widehat S_x\in\mathcal S_x\},\qquad Y_x=A_xE_x.
 \label{eq:support-scoring}
\end{equation}
The primary evaluator uses the released annotations. Additional valid chains found in a separately sampled audit are reported on their own, leaving the primary score untouched. Abstaining on an answerable primary question sets $A_x=0$; on an edited diagnostic item with insufficient evidence, abstaining is the correct response.

\paragraph{Partitioning support errors.}
Among answer-correct but support-incorrect outputs, a missing-only set is a proper subset of a gold set and an extra-only set is a strict superset. All remaining mismatches, including the replacement of a necessary identifier by an irrelevant one, form the wrong-support category. The ordered rule makes the categories disjoint, so the three error rates sum exactly to answer EM minus JEM before rounding. Claim-level unsupported citation rate (UCR) divides unsupported emitted claims by all emitted claims within each seed and then averages over seeds, which differs from the output-level incidence in the human audit, where any unsupported material flags an output.

\subsection{Collecting interactions and updating the agent}
\label{app:training}
The backbone is Qwen2.5-7B-Instruct \citep{yang2025qwen}. Eight seed-specific supervised warm starts use the same 8,000 training-only demonstrations with paired data shuffles, and every method within a seed starts from the same warm-start weights; the SFT row evaluates these eight. LoRA has rank 64, scaling 128, and dropout .05 on the attention projection matrices. The context limit is 32,768 tokens, a complete trajectory holds at most 16 actions and 2,048 actor-generated tokens, and a \textsc{Final} action ends the episode.

Collection uses temperature .8, full-support sampling over the grammar-allowed vocabulary, and eight trajectories per question. The action mask is identical when computing old and new log probabilities. Complete-action ratios include every actor-generated token of the action and exclude tool-response tokens. Evaluation uses greedy decoding and a fixed retrieval index. The eight seeds are 11, 23, 37, 41, 53, 67, 79, and 97.

Training runs on eight A100 80-GB accelerators in bfloat16 with activation checkpointing. Each outer batch holds 64 questions and 512 factual trajectories, with gradient accumulation over eight microbatches. The actor uses AdamW at learning rate $8\times10^{-7}$ with $(\beta_1,\beta_2)=(.9,.95)$, weight decay .1, 3\% warm-up, cosine decay, and gradient-norm clipping at 1.0. PPO clipping is .2 and one proximal epoch follows each collection batch, with the calibrated advantage of Equation~\ref{eq:advantage}. A lagged RMS scale fixed before the batch rescales the optimizer step. Cost baselines use separate regression heads, with expected-cost thresholds of .08 for unsupported claims and .01 for reserved-field access; the dual variables start at zero and follow projected ascent with step .05 and an upper cap of 10.

Every main RL run stops when its generated-token ledger reaches $10^8$ tokens, the last complete batch is retained, and the cost ledger records the small boundary excess. Checkpoint comparison applies the same interpolation convention for that excess across methods, and the primary endpoint is the first full checkpoint at the assigned allowance. Tuning selects hyperparameters, not the final checkpoint. Failed jobs keep their identity and diagnostic record; a numerical interruption is restarted once from its last finite checkpoint with the same seed and configuration, and a second interruption counts as failure. The stability experiment uses its own 32-run roster.

\paragraph{Predictor and allocation.}
A frozen actor representation, action type, text length, retrieval rank, and prefix-visible verifier features feed a two-layer width-256 predictor with output $\tanh(z)$. Separate softplus heads predict the residual second moment and the pair-token cost. Heads use AdamW at $3\times10^{-5}$ and train on the latest four completed batches with weights $1,.7,.49,.343$, and Equation~\ref{eq:critic-loss} supplies the inverse-inclusion loss weights. Predictions for a batch are frozen before its replay jobs begin. The first batch uses zero credit prediction, unit residual weights, and length-based cost prediction.

For vocabulary softmax output $p_\ell$ and sampled token $y_\ell$, the final-layer score is the outer product $(e_{y_\ell}-p_\ell)z_\ell^\top$, and the allocation proxy sums its squared Frobenius norm across actor tokens,
\begin{equation}
 \widetilde G_i=\sum_{\ell\in a_i}\|e_{y_\ell}-p_\ell\|_2^2\|z_\ell\|_2^2.
\end{equation}
Combining this with the predicted residual second moment gives $\widetilde w_i$. The full-action score includes cross-token terms that the proxy drops, which is the price of not materializing a per-action gradient for the entire model. The exact-reference study evaluates allocations with known score vectors, while the actor experiment uses this proxy.

The default lower inclusion bound is .05. Pair-token allowance is $.75F$ for $F$ factual tokens, with proposal tokens accounted separately. Bisection stops when the predicted cost differs from the allowance by less than $10^{-6}$ relatively or after 80 iterations. Selected jobs run to completion, and positive stochastic overshoot is recorded rather than cancelled after seeing returns. A reference proposal uses only the prefix, with at most two schema-validity retries and a valid abstention action as fallback; its distribution is fixed during an outer iteration. Common random streams couple corresponding continuation draws, and the independent-stream variant changes only this coupling.

\subsection{Implementing the comparison methods}
\label{app:baselines}
All comparisons share the actor backbone, warm start, tokenizer, tool interface, task partitions, support evaluator, and generation ledger. The adaptations below define the evaluated systems.

\paragraph{Outcome and retrieval controls.}
PPO-RLVR uses terminal $Y$ with a learned history value, and GRPO centers the same terminal score within the eight trajectories from each question \citep{schulman2017ppo,shao2024deepseekmath}. Search-R1 retains iterative retrieval and masks retrieved tokens out of the actor loss, with its search API mapped to the local context service \citep{jin2025search}. CM-RL uses the same two cost constraints as PCCC-RL and spends its entire allowance on factual trajectories. None of these controls acquires branch labels.

\paragraph{Structured-credit methods.}
TEMPO retains exact token-prefix sharing, branch-gated corrections, and telescoping values, building its response trie from the eight factual rollouts \citep{tran2026tempo}; tool responses are stored as observations, branches with different tool states remain distinct, and the correction coefficient is .5. IBPO retains comparisons across competing reasoning paths and its implicit advantage construction, with eight trajectories supplying the within-question comparisons and a preference temperature of .1 \citep{ding2026ibpo}. Both use the common $Y$ as terminal evaluator, and all rollout tokens are counted.

C3 retains replay from a shared interaction trace and direct counterfactual return credit \citep{chen2026c3}. The local adaptation treats the single actor's sequential tool decisions as trace positions, uses four reference completions at selected positions, and assigns a leave-one-out direct-return baseline, with references drawn uniformly from the same legal prefix-only proposal pool. Multi-agent role scheduling is removed because the interface has one actor. C3 does not use the lagged residual augmentation of Equation~\ref{eq:estimator}. In the identical-information experiment, the C3-style direct update consumes the one-pair bank directly and keeps its return-difference update while matching branch count exactly.

\paragraph{Information controls.}
AM-CF has the same online replay access, legal proposals, predicted inclusion schedule, predictor architecture, and branch-token allowance. It fits a process score to the observed branch contrasts and uses the fitted score as the advantage, without inverse-probability correction at update time, and its trajectories can diverge from ours after the first update. IM-CF freezes the collection checkpoint and receives exactly the same factual histories, actions, selection indicators, probabilities, branch outcomes, and lagged predictions as the calibrated updater, both performing one epoch with the same optimizer. A direct replay row uses $S_iD_i/q_i$, and a shuffled-credit row permutes complete paired contrasts within action type so that their marginal distribution is preserved.

Every method receives a six-configuration tuning allowance: actor learning rate in $\{4,8,12\}\times10^{-7}$ crossed with clipping in $\{.1,.2\}$. The selected rates are $8\times10^{-7}$ for PPO, GRPO, CM-RL, AM-CF, and PCCC-RL, $4\times10^{-7}$ for TEMPO and C3, and $12\times10^{-7}$ for Search-R1 and IBPO. Clipping is .2 except .1 for C3. Method-specific settings above are fixed before this grid. Tuning uses the same validation-token allowance, and each selected configuration is then evaluated under all eight seeds.

\subsection{Computing exact credit and sampled reference values}
\label{app:oracle}
EvidenceSwitch contains 2,000 prefixes with three legal actions and three subsequent three-way decisions, and stores leaf terminal probabilities explicitly. Summation over the $3^3$ future action paths gives $Q(h,a)$ exactly; a prefix-dependent three-action distribution gives $B(h)$, and subtraction gives $\Delta(h,a)$. Prefixes vary evidence necessity, redundant routes, and distractor compatibility. Policy score vectors are computed from the categorical actor, and the oracle record stores leaf values, continuation distributions, reference probabilities, score norms, and exact contrasts.

Each estimator receives 256 independent replay designs per prefix. Within a design, paired returns use a common-uniform coupling with the prescribed marginal distributions, and the model-only, uncorrected, IPS, uniform-augmented, and adaptive-augmented variants share these potential branch returns. Uniform and adaptive designs have the same expected pair cost. Reported bias and RMSE use each prefix's mean estimated credit across the 256 designs, whereas gradient relative mean-squared error uses a complete 2,000-prefix gradient estimate from each single design and averages its squared error over repetitions. The two therefore separate repeated-reference fidelity from the noise of one actor batch.

The QA reference analysis has 64 prefixes per task and 4,096 independent pairs per reference. It compares a separate 256-design low-budget mean against the reference and reports that reference's Monte Carlo standard error, and noise-adjusted squared error subtracts the average reference variance before the square root is taken. Restoration auditing compares pre-action counters, observation buffers, identifiers, and the next observation under identical random streams, with denominators reported explicitly.

\subsection{Language edits, citation transfer, and human review}
\label{app:language}
Six edit families have 300 unique original--edited pairs each, and every pair is evaluated by all eight trained seed models, giving 2,400 binary outcomes per family and method. Paraphrase and irrelevant insertion require the same supported answer on both inputs. Evidence deletion requires either a supported alternative chain or an evidence-insufficient response. Relation, scope, and value edits require the response and its cited span to track the changed proposition. The four meaning-changing families are averaged with equal weight. Edits are local, source identifiers persist, and the edited gold response is fixed before method outputs are compared.

ALCE-ASQA uses 948 questions and the common retrieval bundle \citep{gao2023alce}. STR-EM is the mean fraction of accepted short-answer facets covered by the long answer. Citation precision is the proportion of attached references that support their associated claim, and citation recall is the proportion of claims with an entailing citation; both are macro-averaged over outputs and then over seeds, and an answer with no citation scores zero on each. Output-token length, references per output, and omitted facets accompany these quality scores.

The blind audit samples 200 prompts from each of HotpotQA, MuSiQue, and ASQA, with 25 prompts per seed within task, and all three methods answer the same prompts. Three raters independently label each output for any unsupported claim, contradictory citation, or evidence-insufficient answer presented as established fact, with method names and formatting cues hidden. The primary incidence uses majority vote, while Fleiss' $\kappa$, pairwise agreement, and unanimity use the pre-adjudication ratings; the roster holds $600\times3\times3=5,400$ judgments. A separate alternative-chain review inspects 120 answer-correct outputs with incorrect support per method and task, checks the complete proposed chain, and records an adjudicated validity decision.

\subsection{Measuring computation and update stability}
\label{app:compute}
The generation ledger separates factual actor tokens, proposal tokens, and both continuation branches. Attention, forward, backward, auxiliary-head, and verifier operations all enter total FLOPs. Peak memory is measured per accelerator, accelerator-hours sum allocated device-seconds over devices, and wall time equals accelerator-hours divided by eight plus serialized retrieval, restoration, and verifier service time. The token-matched endpoint, the fixed-factual-token overhead experiment, and the resource-indexed frontier are three distinct comparisons.

The fixed-factual experiment uses 50 million factual tokens per seed, with adaptive and uniform replay sharing their nominal pair allowance; predicted-length error and differing completed branches determine realised consumption, and the four-pair variant has an expanded allowance. Frontier curves retain every saved checkpoint, including regressions, and the common FLOP grid is linearly interpolated between adjacent checkpoint observations within each seed. Endpoint hypothesis tests use the directly evaluated primary checkpoints.

The floor study uses 16,000 candidate decisions and 2,500,746 factual tokens. Floors .02, .05, and .10 share $B=.75F$, while the .20 stress setting has $B=1.50F$ and is displayed with that distinct allowance. Logs include predicted and realised pair cost, proposal cost, all inclusion probabilities, residuals, calibrated advantages, and finite-update action ratios. Variance is the conditional design variance of the score-weighted residual correction, normalized to the .05 row. The clipping fraction counts complete-action ratios outside $[.8,1.2]$, and KL uses the full categorical action distribution. The separate 32-run stability roster uses 24 independently resampled 128-decision minibatches per restart; an update is rejected when mean action KL exceeds .025, two consecutive rejections stop that restart, every KL is computed from the complete old and proposed distributions, and the ledger keeps all rejected updates. The policy-drift grid crosses refresh intervals $1,2,4,8$ with actor step multipliers $.5,1,2$ and evaluates 1,200 common prefixes per configuration.

\section{Statistical Analysis and Numerical Aggregation}
\label{app:statistics}
Repeated runs share questions and evidence sources, so their outcomes are not all independent observations. This appendix defines the quantity compared between methods and the sampling units used to quantify uncertainty. The paired tests assess differences across runs, while bootstrap intervals describe variation across seeds and evidence sources.
\subsection{What each paired comparison estimates}
The primary quantity of interest, or estimand, is the difference in JEM between two methods on the locked evaluation questions, averaged over their paired training seeds. Let $Y_{msi}$ denote the binary joint outcome for method $m$, seed $s$, and question $i$. For two methods, the paired seed effect in percentage points is
\begin{equation}
 d_s=\frac{100}{N}\sum_{i=1}^N(Y_{1si}-Y_{0si}),\qquad
 \widehat\tau=\frac18\sum_{s=1}^8d_s.
\end{equation}
The six-member primary family comprises four online comparisons and two single-update identical-information comparisons. The latter contrast changes in JEM from a shared checkpoint, not unrelated final models. All six effects, confidence intervals, raw probabilities, and Holm-adjusted probabilities appear in Table~\ref{tab:inference}.

\paragraph{Exact paired test.}
Under the paired null, the seed-level difference vector is sign-symmetric, with independent seed randomization. Enumerating $\varepsilon\in\{-1,1\}^8$ gives
\begin{equation}
 p=\frac{1}{2^8}\sum_{\varepsilon}
 \ind\left\{\left|\frac18\sum_s\varepsilon_sd_s\right|
 \ge |\widehat\tau|\right\}.
 \label{eq:signflip}
\end{equation}
Ties are included. All 256 assignments are enumerated rather than approximated by repeated Monte Carlo permutations. When all eight nonzero differences have the same sign, the raw probability is exactly $2/256=.0078125$. Multiple comparisons can therefore have identical raw probabilities despite different effect magnitudes and dispersions.

Sort the six probabilities as $p_{(1)}\le\cdots\le p_{(6)}$. Holm adjustment is
\begin{equation}
 p^{\mathrm H}_{(j)}=\min\left\{1,\max_{k\le j}(7-k)p_{(k)}\right\}.
\end{equation}
The four online contrasts share the minimum attainable raw probability and have adjusted value .046875. The two identical-information contrasts retain their larger adjusted probabilities. Table~\ref{tab:seeds} gives the underlying online seed means. Its rounded entries are descriptive, and inference uses the full-precision binary record.

\begin{table*}[!t]
\centering\tabsetup
\small
\begin{tabular}{llrrrrr}
\toprule
Task & Contrast & $\Delta$ & 95\% CI & $p$ & $p_{\mathrm{Holm}}$ & Positive\\
\midrule
HotpotQA & PCCC--CM-RL & 4.88 & [3.99, 5.76] & 0.0078125 & 0.0468750 & 8/8 \\
HotpotQA & PCCC--AM-CF & 4.19 & [2.87, 5.58] & 0.0078125 & 0.0468750 & 8/8 \\
MuSiQue & PCCC--CM-RL & 5.80 & [4.26, 7.41] & 0.0078125 & 0.0468750 & 8/8 \\
MuSiQue & PCCC--AM-CF & 3.86 & [2.63, 5.13] & 0.0078125 & 0.0468750 & 8/8 \\
HotpotQA & Bank: calibrated--IM & 0.12 & [-0.17, 0.41] & 0.3281250 & 0.3281250 & 6/8 \\
MuSiQue & Bank: calibrated--IM & 0.36 & [-0.09, 0.79] & 0.1093750 & 0.2187500 & 5/8 \\
\bottomrule
\end{tabular}
\caption{All six primary tests. Effects and intervals are JEM percentage points. Exact two-sided sign flips enumerate 256 seed sign patterns. Intervals use 10,000 paired seed--source-cluster bootstrap replicates. Holm correction uses the six rows as one family. The online and one-update bank contrasts have distinct estimands.}
\label{tab:inference}
\end{table*}

\begin{table*}[!t]
\centering\tabsetup
\small
\begin{tabular}{lrrrrrr}
\toprule
Seed & \multicolumn{3}{c}{HotpotQA} & \multicolumn{3}{c}{MuSiQue}\\
 & CM-RL & AM-CF & PCCC-RL & CM-RL & AM-CF & PCCC-RL\\
\midrule
11 & 53.22 & 51.94 & 58.50 & 29.75 & 31.44 & 34.88 \\
23 & 52.91 & 54.11 & 55.85 & 31.57 & 32.35 & 35.50 \\
37 & 52.67 & 52.72 & 57.85 & 32.52 & 36.53 & 38.35 \\
41 & 50.79 & 50.29 & 57.22 & 30.49 & 31.53 & 37.36 \\
53 & 50.71 & 53.45 & 56.30 & 33.93 & 32.40 & 37.03 \\
67 & 51.15 & 52.83 & 55.77 & 30.24 & 32.69 & 36.41 \\
79 & 51.21 & 52.55 & 55.67 & 31.77 & 33.88 & 37.90 \\
97 & 52.01 & 52.26 & 56.52 & 27.93 & 32.89 & 37.15 \\
Mean & 51.83 & 52.52 & 56.71 & 31.03 & 32.96 & 36.82 \\
SD & 1.00 & 1.13 & 1.05 & 1.84 & 1.64 & 1.17 \\
\bottomrule
\end{tabular}
\caption{Paired seed results (JEM, \%). Each seed uses 7,405 HotpotQA and 2,417 MuSiQue items. Unrounded integer successes determine the means, SDs, and tests. Two-decimal display rounding is applied last.}
\label{tab:seeds}
\end{table*}

\subsection{Resampling seeds and shared evidence sources}
Questions sharing an evidence source can have dependent errors. Let $c(i)$ be the connected source-document component of question $i$, with $C$ components. Each bootstrap draw independently samples seed multiplicities $u_s$ from a multinomial with eight draws and component multiplicities $v_c$ from a multinomial with $C$ draws. It preserves every method pair and every question within a sampled component. The replicate effect is
\begin{equation}
 \widehat\tau^*=100\frac{\sum_{s,i}u_sv_{c(i)}(Y_{1si}-Y_{0si})}
 {\sum_{s,i}u_sv_{c(i)}}.
 \label{eq:crossed}
\end{equation}
The denominator retains the sizes of unequal components. The 2.5th and 97.5th percentiles of 10,000 replicates form the main interval. Prespecified subgroup curves use 3,000 replicates per stratum with the same pairing. This crossed resampling follows the two-factor dependence structure \citep{owen2007}. Primary tests and intervals answer different randomization and sampling questions.

For a factorial contrast, the per-question contribution is formed from all four paired cells before resampling. Writing $\bar Y_{ck}$ for the mean at credit setting $c$ and constraint setting $k$, the effects are
\begin{align}
 \tau_C&=\tfrac12[(\bar Y_{10}-\bar Y_{00})+(\bar Y_{11}-\bar Y_{01})],\\
 \tau_K&=\tfrac12[(\bar Y_{01}-\bar Y_{00})+(\bar Y_{11}-\bar Y_{10})],\\
 \tau_{CK}&=\bar Y_{11}-\bar Y_{10}-\bar Y_{01}+\bar Y_{00}.
\end{align}
The same definitions apply to seed-level UCR. Contrasts are computed in floating point before aggregation. All tables round only their displayed values.

\subsection{Counting repeated tests and rater agreement}
A single seed's language-family success is $100k_s/300$, where $k_s$ is an integer. Averaging the eight seed rates is exactly $100\sum_sk_s/2400$. Thus a rate such as $2213/2400=92.2083\%$ has a valid integer denominator while the number of unique language pairs remains 300. Language-table uncertainty is the sample SD of eight seed rates. The 2,400 evaluations do not replace the 300 item identities in resampling. The four-family response score divides total success count by 9,600.

For three ratings on output $i$, let $n_{ij}$ be the number assigned to binary category $j$. With $n=3$ raters and $M$ outputs,
\begin{align}
 P_i&=\frac{\sum_j n_{ij}(n_{ij}-1)}{n(n-1)},\\
 p_j&=\frac{\sum_i n_{ij}}{Mn},\\
 \kappa&=\frac{\bar P-\sum_jp_j^2}{1-\sum_jp_j^2}.
 \label{eq:fleiss}
\end{align}
Pairwise agreement is $100\bar P$. Unanimous agreement counts outputs with three identical labels. The majority-vote incidence and $\kappa$ intervals resample prompts within each task 10,000 times, keeping all three ratings and all method outputs for a prompt together. Task strata retain their 200-prompt weights. The 600-output aggregate is not obtained by treating 1,800 ratings as independent outputs.

Zero restoration mismatches are accompanied by their tested sample counts. With $n$ independent audit draws and no mismatch, the one-sided exact 95\% upper rate is $1-.05^{1/n}$. This calculation is used only for the restoration check specified in the audit. Noise-corruption sweeps preserve the clean evaluation scorer, so corrupted training labels and measured test performance are separate variables.

\subsection{From recorded outcomes to displayed values}
Primary outcome records contain task, method, seed, item and source identifiers, answer EM, support EM, JEM, answer F1, support-error category, and claim counts. The single-update bank additionally contains pre-update and post-update JEM. The language record stores edit family, original--edited pair identity, seed, method, and binary success. The rating record stores output identity, task, method, seed, rater slot, and label. Cost records separate resources by method, seed, and checkpoint.

The analysis pipeline computes seed means, counts, uncertainty, tests, and tables from these records. Scalar references in the abstract and results use the same generated macros as the tables. Default conditions in ablation, scale, noise, and floor analyses reuse the primary condition rather than copying rounded values. Learning curves share the same SFT origin within each seed and reuse the primary binary outcomes at the primary endpoint. Hashes in the manifest identify the distributed analysis files and permit exact reconstruction of the figures and PDF.
\FloatBarrier

\section{Extended Results and Mechanism Analysis}
\label{app:results}
The main text follows one path, from measuring a decision to improving answers, responding to changed evidence, and accounting for compute. This appendix supplies the additional controls behind those claims, and every table keeps the full evaluated conditions, reversals and uncertain effects included.
\begin{table*}[!t]
\centering\tabsetup
\small
\begin{tabular}{lrrrrrr}
\toprule
Estimator & Bias & RMSE & NRMSE & $\rho$ & Sign (\%) & Grad. rMSE\\
\midrule
Model only & 0.0353 & 0.1160 & 0.586 & 0.811 & 83.61 & 0.0984 \\
No residual correction & 0.0266 & 0.0922 & 0.465 & 0.881 & 88.05 & 0.0564 \\
IPS only & -0.0027 & 0.0690 & 0.348 & 0.945 & 91.92 & 0.0739 \\
Uniform augmented & 0.0002 & 0.0649 & 0.327 & 0.952 & 95.28 & 0.0733 \\
Adaptive augmented & -0.0037 & 0.0666 & 0.336 & 0.944 & 94.37 & 0.0682 \\
\bottomrule
\end{tabular}
\caption{Exact-reference evaluation on 2,000 independent finite-tree prefixes. Credit bias, RMSE, rank correlation, and sign use the per-prefix mean over 256 independent replay repetitions. NRMSE divides by the SD of exact credit. Gradient rMSE averages squared error of one 2,000-prefix gradient batch across repetitions and normalizes by squared exact-gradient norm. The allocation target is gradient variance.}
\label{tab:oracle}
\end{table*}

\begin{table*}[!t]
\centering\tabsetup
\small
\begin{tabular}{lrrrrrr}
\toprule
Task & Prefixes & Median SE & P95 SE & RMSE & Noise-adj. RMSE & $\rho$\\
\midrule
HotpotQA & 64 & 0.0069 & 0.0081 & 0.0753 & 0.0750 & 0.941 \\
MuSiQue & 64 & 0.0068 & 0.0081 & 0.0576 & 0.0572 & 0.956 \\
\bottomrule
\end{tabular}
\caption{High-budget QA credit reference: 4,096 independent pairs at each of 64 prefixes per task. The compared estimate uses 256 independent randomized-replay designs per prefix. Noise-adjusted RMSE subtracts mean reference-mean variance from squared error before taking the square root.}
\label{tab:reference}
\end{table*}

\subsection{The identical-information and exact-credit controls}
Table~\ref{tab:bank} expands the identical-information comparison. Direct replay produces the largest mean one-update gain on MuSiQue, while calibrated replay has the strongest primary online endpoint. At a single checkpoint the return differences, prediction residuals, and optimizer step interact differently from repeated recollection, which is our reading of why the two orderings disagree. The shuffled control preserves the distribution of contrast values while breaking their assignment to the relevant action, and its near-zero mean gains make the association visible without changing the quantity of replay information at all.

Table~\ref{tab:oracle} reports repeated-design credit fidelity and single-design gradient error side by side. Uniform augmentation has the best mean credit RMSE and adaptive augmentation the lower gradient error. The uncorrected estimator's shrinkage yields a lower finite-sample gradient error at this particular oracle distribution despite its positive credit bias, a reminder that bias and usefulness are separate quantities and that we report both rather than folding them into one ranking.

\begin{table}[H]
\centering\tabsetup
\small
\begin{tabular}{lrr}
\toprule
Update rule & HotpotQA gain & MuSiQue gain\\
\midrule
Terminal update & $0.29\,\pm\,0.32$ & $0.16\,\pm\,0.42$ \\
IM-CF & $0.45\,\pm\,0.31$ & $0.18\,\pm\,0.45$ \\
Direct replay & $0.57\,\pm\,0.26$ & $0.82\,\pm\,0.32$ \\
Calibrated & $0.57\,\pm\,0.24$ & $0.53\,\pm\,0.48$ \\
Shuffled pairs & $0.03\,\pm\,0.34$ & $0.16\,\pm\,0.18$ \\
\bottomrule
\end{tabular}
\caption{Complete identical-bank update control (JEM points, eight-seed mean $\pm$ SD). The direct-replay row uses the C3-style observed contrast. The shuffled row permutes residual--decision associations within action type.}
\label{tab:bank}
\end{table}

\begin{table}[H]
\centering\tabsetup
\small
\begin{tabular}{lrrr}
\toprule
Task & Restorations & Mismatches & Upper 95\% (\%)\\
\midrule
HotpotQA & 20480 & 0 & 0.0146 \\
MuSiQue & 19456 & 0 & 0.0154 \\
\bottomrule
\end{tabular}
\caption{Exact replay-state audit. A mismatch means a byte-level difference in any restored state field or a changed next observation under the same action and RNG state. The last column is the one-sided exact binomial upper endpoint for the mismatch rate.}
\label{tab:restoration}
\end{table}

The high-budget reference carries its own Monte Carlo uncertainty (Table~\ref{tab:reference}), and noise adjustment changes the RMSE only slightly because the 4,096-pair reference has a much smaller standard error than the lower-budget estimate. The restoration test observes zero mismatches in the specified state check, and Table~\ref{tab:restoration} gives the denominator together with the exact upper rate implied by it. The finite-tree calculation also stores the full leaf array and policy transitions, so the reference contrast can be recomputed by direct summation at any time.

\begin{table*}[!t]
\centering\tabsetup
\small
\begin{tabular}{llrrrrr}
\toprule
Task & Stratum & Items & CM-RL & AM-CF & PCCC-RL & $\Delta_{\rm AM}$ [95\% CI]\\
\midrule
HotpotQA & Hops 2 & 7405 & 51.83 & 52.52 & 56.71 & 4.19 [2.84, 5.54] \\
HotpotQA & Unique route & 4587 & 48.57 & 49.22 & 54.27 & 5.05 [3.58, 6.68] \\
HotpotQA & Redundant & 2818 & 57.15 & 57.89 & 60.69 & 2.80 [1.27, 4.28] \\
MuSiQue & Hops 2 & 1194 & 35.29 & 37.43 & 40.53 & 3.10 [1.23, 4.99] \\
MuSiQue & Hops 3 & 784 & 29.66 & 31.33 & 35.87 & 4.54 [2.17, 6.85] \\
MuSiQue & Hops 4 & 439 & 21.87 & 23.75 & 28.45 & 4.70 [2.32, 7.28] \\
MuSiQue & Unique route & 1958 & 29.85 & 31.69 & 36.04 & 4.35 [3.16, 5.63] \\
MuSiQue & Redundant & 459 & 36.03 & 38.40 & 40.14 & 1.74 [-1.31, 4.73] \\
2Wiki & Hops 2 & 8664 & 54.52 & 54.32 & 55.80 & 1.48 [0.31, 2.73] \\
2Wiki & Hops 3 & 2160 & 44.43 & 45.01 & 47.42 & 2.41 [1.06, 3.87] \\
2Wiki & Hops 4 & 1752 & 37.06 & 37.38 & 40.09 & 2.71 [0.50, 4.95] \\
2Wiki & Unique route & 6852 & 45.89 & 46.51 & 49.49 & 2.98 [1.58, 4.34] \\
2Wiki & Redundant & 5724 & 55.71 & 54.97 & 55.38 & 0.41 [-0.69, 1.68] \\
\bottomrule
\end{tabular}
\caption{Reasoning structure and evidence redundancy. Within each task, the hop rows and the route rows are separate complete partitions. Gains compare paired PCCC-RL and AM-CF predictions. Intervals resample seeds and source clusters within the stratum.}
\label{tab:strata}
\end{table*}

\subsection{Separating credit, constraints, and action types}
\begin{table}[H]
\centering\tabsetup
\small
\begin{tabular}{llrr}
\toprule
Credit & Costs & JEM & UCR\\
\midrule
Off & Off & $30.14\,\pm\,1.83$ & $17.30\,\pm\,0.28$ \\
On & Off & $36.15\,\pm\,1.45$ & $14.05\,\pm\,0.39$ \\
Off & On & $31.03\,\pm\,1.84$ & $9.59\,\pm\,0.79$ \\
On & On & $36.82\,\pm\,1.17$ & $6.28\,\pm\,0.48$ \\
\bottomrule
\end{tabular}
\caption{Credit $\times$ constraints on MuSiQue. JEM and claim-level UCR are eight-seed means $\pm$ SD in percent. All four cells use the same token budget.}
\label{tab:factorial}
\end{table}

The four cells independently enable decision credit and cost constraints. Their interaction is \metric{factor_interaction} (\metric{factor_interaction_ci}), while constraints reduce unsupported claims by \metric{factor_cost_ucr} points against \metric{factor_credit_ucr} for credit. That asymmetry is what the controls are for: a better learning signal and a stronger cost penalty can produce similar JEM, and only this design tells them apart.

\begin{table}[H]
\centering\tabsetup
\small
\begin{tabular}{lrr}
\toprule
Variant & MuSiQue JEM & $\Delta$ to full\\
\midrule
PCCC-RL & $36.82\,\pm\,1.17$ & 0.00 \\
Uniform replay & $35.86\,\pm\,1.44$ & -0.97 \\
Model only & $34.53\,\pm\,1.40$ & -2.29 \\
IPS only & $35.03\,\pm\,0.84$ & -1.79 \\
No residual correction & $35.12\,\pm\,0.82$ & -1.70 \\
Independent branches & $36.66\,\pm\,1.24$ & -0.16 \\
Action-dependent reference & $35.32\,\pm\,1.19$ & -1.50 \\
Credit added twice & $35.85\,\pm\,1.09$ & -0.97 \\
Fixed q: weak predictor & $35.12\,\pm\,1.56$ & -1.71 \\
Fixed q: zero predictor & $35.22\,\pm\,1.38$ & -1.60 \\
Remove SEARCH credit & $35.64\,\pm\,0.88$ & -1.18 \\
Remove READ credit & $36.14\,\pm\,1.73$ & -0.68 \\
Remove VERIFY credit & $35.73\,\pm\,1.77$ & -1.10 \\
Remove FINAL credit & $36.13\,\pm\,1.31$ & -0.69 \\
\bottomrule
\end{tabular}
\caption{Policy-learning ablations: eight paired seeds at $10^8$ generated tokens. The first five rows fix the prefix-only reference family. The fixed-$q$ rows hold the entire inclusion vector and reference bank fixed while changing the predictor. The action-removal rows use terminal advantages at the named action type and calibrated credit elsewhere.}
\label{tab:ablations}
\end{table}

Table~\ref{tab:ablations} keeps the legal prefix-only reference family fixed across the model-only, IPS, uncorrected, and augmented estimates. The fixed-$q$ variants freeze the complete inclusion schedule and reference bank while replacing the predictor with a weakened or zero predictor, and their JEM values stay close, with the weakened predictor slightly below zero prediction. Predictive fit and update utility are therefore evaluated separately. Independent future streams change pair covariance, action-dependent references change the gradient baseline, and adding credit twice changes the actor objective rather than its observation budget; each of the three is a different failure and each shows up as a different drop.

For action removal, terminal advantages replace calibrated credit only at the named decision type. Search and verification account for larger losses than reading or finalization, and Table~\ref{tab:actions} records action counts and exact repeated calls so that these comparisons can be read against actual trajectory length, which stays close across methods. The support-error decomposition in Table~\ref{tab:errors} locates where the improved JEM comes from among answer-correct trajectories.
\begin{table}[H]
\centering\tabsetup
\setlength{\tabcolsep}{2.5pt}
\small
\begin{tabular}{lrrr}
\toprule
Method & \shortstack{Actions per\\episode} & \shortstack{Repeated\\(\%)} & \shortstack{Cap-hit\\counts}\\
\midrule
\multicolumn{4}{l}{\textbf{HotpotQA}} \\
CM-RL & 6.21 & 9.63 & 145/59240 \\
AM-CF & 6.20 & 8.09 & 109/59240 \\
C3 & 6.22 & 7.11 & 141/59240 \\
PCCC-RL & 5.91 & 5.98 & 78/59240 \\
\multicolumn{4}{l}{\textbf{MuSiQue}} \\
CM-RL & 6.90 & 9.51 & 55/19336 \\
AM-CF & 6.89 & 8.25 & 56/19336 \\
C3 & 6.89 & 7.07 & 42/19336 \\
PCCC-RL & 6.59 & 6.00 & 29/19336 \\
\bottomrule
\end{tabular}
\caption{Trajectory length and repetition at the primary endpoints. Repeated actions have identical tool name and normalized arguments. Every completed, failed, and cap-hit episode stays in the denominator.}
\label{tab:actions}
\end{table}

\begin{table}[H]
\centering\tabsetup
\setlength{\tabcolsep}{2pt}
\small
\begin{tabular}{lrrrrrr}
\toprule
Method & \shortstack{Answer\\EM} & JEM & \shortstack{Support\\errors} & Missing & Extra & Wrong\\
\midrule
\multicolumn{7}{l}{\textbf{HotpotQA}} \\
CM-RL & 70.51 & 51.83 & 18.68 & 10.88 & 4.66 & 3.14 \\
AM-CF & 72.07 & 52.52 & 19.56 & 11.05 & 4.97 & 3.54 \\
C3 & 73.62 & 55.82 & 17.80 & 10.17 & 4.43 & 3.20 \\
PCCC-RL & 72.84 & 56.71 & 16.13 & 7.76 & 4.05 & 4.32 \\
\multicolumn{7}{l}{\textbf{MuSiQue}} \\
CM-RL & 49.62 & 31.03 & 18.60 & 11.31 & 4.63 & 2.66 \\
AM-CF & 51.20 & 32.96 & 18.24 & 11.08 & 4.51 & 2.64 \\
C3 & 54.39 & 35.93 & 18.46 & 10.95 & 4.75 & 2.76 \\
PCCC-RL & 54.37 & 36.82 & 17.55 & 9.36 & 4.13 & 4.06 \\
\multicolumn{7}{l}{\textbf{2Wiki}} \\
CM-RL & 69.93 & 50.36 & 19.58 & 11.11 & 4.97 & 3.50 \\
AM-CF & 69.96 & 50.36 & 19.59 & 11.19 & 4.89 & 3.51 \\
C3 & 70.89 & 52.38 & 18.51 & 10.47 & 4.67 & 3.37 \\
PCCC-RL & 70.52 & 52.17 & 18.35 & 8.67 & 4.69 & 4.99 \\
\bottomrule
\end{tabular}
\caption{Support-error decomposition (percent of all seed--item evaluations). Missing, extra, and wrong identifiers form a disjoint partition of the support-error column among answer-correct outputs, with precedence wrong $>$ extra $>$ missing when several faults coexist. \shortstack{Answer\\EM} equals JEM plus that column.}
\label{tab:errors}
\end{table}

\begin{table*}[!t]
\centering\tabsetup
\small
\begin{tabular}{lrrrr}
\toprule
Edit family & CM-RL & AM-CF & C3 & PCCC-RL\\
\midrule
Paraphrase & \shortstack{85.58$\pm$2.04\\2054/2400} & \shortstack{89.21$\pm$2.23\\2141/2400} & \shortstack{90.00$\pm$1.58\\2160/2400} & \shortstack{92.21$\pm$1.38\\2213/2400} \\
Irrelevant insertion & \shortstack{91.04$\pm$1.69\\2185/2400} & \shortstack{93.08$\pm$1.50\\2234/2400} & \shortstack{92.50$\pm$1.27\\2220/2400} & \shortstack{93.83$\pm$1.10\\2252/2400} \\
Evidence deletion & \shortstack{63.46$\pm$2.82\\1523/2400} & \shortstack{68.67$\pm$2.68\\1648/2400} & \shortstack{73.17$\pm$1.80\\1756/2400} & \shortstack{74.83$\pm$2.38\\1796/2400} \\
Relation & \shortstack{58.83$\pm$2.48\\1412/2400} & \shortstack{65.83$\pm$3.91\\1580/2400} & \shortstack{67.42$\pm$2.97\\1618/2400} & \shortstack{69.38$\pm$2.48\\1665/2400} \\
Scope & \shortstack{50.58$\pm$4.41\\1214/2400} & \shortstack{55.88$\pm$3.43\\1341/2400} & \shortstack{62.83$\pm$2.64\\1508/2400} & \shortstack{62.25$\pm$3.30\\1494/2400} \\
Value & \shortstack{60.96$\pm$2.13\\1463/2400} & \shortstack{64.75$\pm$0.61\\1554/2400} & \shortstack{66.38$\pm$3.60\\1593/2400} & \shortstack{71.00$\pm$2.13\\1704/2400} \\
\bottomrule
\end{tabular}
\caption{Language-pair outcomes. Each family contains 300 unique pairs evaluated under all eight seeds, hence 2,400 binary outcomes per method and family. Cells show the mean $\pm$ between-seed SD (\%) and integer successes out of 2,400. Repeated evaluations retain pair and source-cluster identity in inference.}
\label{tab:language}
\end{table*}

\subsection{Missing links and alternative evidence chains}
The hop and redundancy rows of Table~\ref{tab:strata} define two separate partitions, each summing to the task's evaluation set, and the intervals retain source-cluster pairing. HotpotQA has two annotated hops, while MuSiQue separates two-, three-, and four-hop composition. A unique evidence route makes the deletion of one necessary premise consequential; a redundant route permits recovery through a second chain. We therefore read the gain jointly with route redundancy rather than attributing it to task name, since the two tasks differ in both at once.

\begin{figure}[H]
\centering\includegraphics[width=\columnwidth]{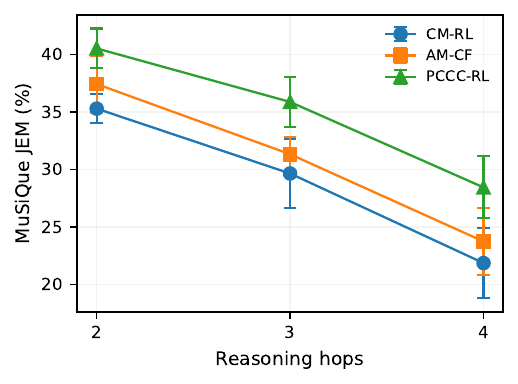}
\caption{MuSiQue JEM by annotated reasoning depth. Each depth group holds the same question identities across methods, error bars are between-seed SD, and subgroup sizes appear in Table~\ref{tab:strata}.}
\label{fig:hops}
\end{figure}

\begin{table}[H]
\centering\tabsetup
\small
\begin{tabular}{llrr}
\toprule
Task & Method & Valid-chain counts & Rate (\%)\\
\midrule
HotpotQA & CM-RL & 18/120 & 15.00 \\
HotpotQA & AM-CF & 17/120 & 14.17 \\
HotpotQA & C3 & 33/120 & 27.50 \\
HotpotQA & PCCC-RL & 30/120 & 25.00 \\
MuSiQue & CM-RL & 11/120 & 9.17 \\
MuSiQue & AM-CF & 21/120 & 17.50 \\
MuSiQue & C3 & 21/120 & 17.50 \\
MuSiQue & PCCC-RL & 27/120 & 22.50 \\
\bottomrule
\end{tabular}
\caption{Alternative-support review among answer-correct, exact-support-mismatched outputs. Each task--method stratum contributes 120 outputs, giving 960 cases. A valid alternative supplies a complete entailment chain using only observed passages.}
\label{tab:alternatives}
\end{table}

The alternative-chain audit samples only answer-correct outputs that fail the released support-set check, so its denominator is the audited subset rather than all questions. Table~\ref{tab:alternatives} gives the valid-chain count and rate for every method. C3 and PCCC-RL both find chains absent from the primary annotation, which complements the support-error partition while leaving the primary benchmark scores unchanged, and it is also a reminder that a JEM gap of a point or two partly reflects annotation coverage rather than behaviour alone.

\begin{table*}[!t]
\centering\tabsetup
\small
\begin{tabular}{lrrrrr}
\toprule
Method & HotpotQA & MuSiQue & ASQA & Overall & Incidence [95\% CI]\\
\midrule
CM-RL & 21/200 (10.5\%) & 17/200 (8.5\%) & 35/200 (17.5\%) & 73/600 & 12.17 [9.67, 14.83] \\
AM-CF & 14/200 (7.0\%) & 20/200 (10.0\%) & 21/200 (10.5\%) & 55/600 & 9.17 [7.00, 11.50] \\
PCCC-RL & 9/200 (4.5\%) & 12/200 (6.0\%) & 25/200 (12.5\%) & 46/600 & 7.67 [5.67, 9.83] \\
\bottomrule
\end{tabular}
\caption{Blinded unsupported-claim audit. There are 600 matched prompts, 1,800 method outputs, and 5,400 independent binary judgments. Task cells count majority-positive outputs. Every task has 200 outputs per method. Overall intervals use 10,000 task-stratified prompt bootstrap replicates.}
\label{tab:audit}
\end{table*}

\subsection{When selective replay produces large or stale corrections}
A selection floor is the smallest allowed replay probability. Lowering it saves checks on some decisions but raises their correction weight whenever they are drawn, which is a sampling-noise effect; staleness is a separate one, arising when the agent has moved since credit was measured. We examine them apart because they call for different remedies.
\begin{table}[H]
\centering\tabsetup
\setlength{\tabcolsep}{3pt}
\small
\begin{tabular}{lrrrr}
\toprule
$q_{\min}$ & 0.02 & 0.05 & 0.10 & 0.20 \\
\midrule
$B/F$ & 0.75 & 0.75 & 0.75 & 1.50 \\
Mean $q$ & 0.138 & 0.137 & 0.132 & 0.263 \\
Pair M & 1.904 & 1.921 & 1.899 & 3.803 \\
Proposal M & 0.330 & 0.334 & 0.318 & 0.635 \\
Util. (\%) & 101.51 & 102.44 & 101.27 & 101.39 \\
Gen.$\times$ & 1.893 & 1.902 & 1.887 & 2.775 \\
JEM & \shortstack{$35.73$\\$\pm\,1.35$} & \shortstack{$36.82$\\$\pm\,1.17$} & \shortstack{$36.91$\\$\pm\,1.17$} & \shortstack{$36.22$\\$\pm\,1.10$} \\
\bottomrule
\end{tabular}
\caption{Budget-accounted floor sensitivity. $F=2{,}500{,}746$ factual tokens for the common diagnostic bank. Expected pair budget is $0.75F$ for the first three settings and $1.50F$ for the 0.20 stress setting. Pair and proposal counts are realized totals in millions. Utilization is realized pair tokens divided by $B$. Generation is $(F+\mathrm{pair}+\mathrm{proposal})/F$. JEM comes from the paired training runs for each design.}
\label{tab:budget}
\end{table}

\begin{table}[H]
\centering\tabsetup
\small
\begin{tabular}{lrrrr}
\toprule
$q_{\min}$ & 0.02 & 0.05 & 0.10 & 0.20 \\
\midrule
$(1/q)_{99}$ & 49.70 & 20.00 & 10.00 & 5.00 \\
$|D-d|_{99}$ & 1.138 & 1.134 & 1.147 & 1.149 \\
$|\hat\Delta|_{99}$ & 6.78 & 6.97 & 8.09 & 4.71 \\
$|\hat\Delta|_{99.9}$ & 16.07 & 16.51 & 10.74 & 5.62 \\
Var.$\times$ & 1.033 & 1.000 & 0.921 & 0.455 \\
Clip (\%) & 5.61 & 5.62 & 5.46 & 6.47 \\
KL & 0.0128 & 0.0125 & 0.0121 & 0.0069 \\
Fail & 2/32 & 4/32 & 3/32 & 0/32 \\
\bottomrule
\end{tabular}
\caption{Tail and finite-update diagnostics for 16,000 candidate decisions. Inverse-probability and advantage quantiles use all candidates. Residual quantiles use replayed pairs. Var.$\times$ is the conditional score-weighted gradient variance relative to the 0.05 design. Clip is the fraction of complete-action ratios outside $[0.8,1.2]$. KL is computed over complete action distributions. Fail counts out of 32 independent stress restarts use the locked instability criterion.}
\label{tab:tails}
\end{table}

\begin{table}[H]
\centering\tabsetup
\small
\begin{tabular}{lrrrr}
\toprule
$q_{\min}$ & 0.02 & 0.05 & 0.10 & 0.20 \\
\midrule
Mean $d$ & 0.0451 & 0.0451 & 0.0451 & 0.0451 \\
Var. $d$ & 0.0267 & 0.0267 & 0.0267 & 0.0267 \\
Mean $\hat\Delta$ & 0.0172 & 0.0108 & 0.0133 & 0.0137 \\
Var. $\hat\Delta$ & 2.0297 & 1.9019 & 1.6355 & 0.8408 \\
$|\hat\Delta|_{95}$ & 1.367 & 1.446 & 1.379 & 1.889 \\
Top 1\% & 26.54 & 25.93 & 24.63 & 15.40 \\
Cost MAPE & 17.30 & 17.30 & 17.30 & 17.30 \\
\bottomrule
\end{tabular}
\caption{Pre- and post-calibration moments on the same 16,000 candidates. The predictor and cost forecasts are held fixed across floors, so their repeated values reflect the shared design. Top 1\% is the fraction of absolute score-weighted advantage mass carried by the largest one percent of decisions. Cost MAPE is the mean absolute relative pair-cost prediction error (\%). The .20 setting uses the expanded budget.}
\label{tab:advantage-moments}
\end{table}

Table~\ref{tab:advantage-moments} gives predictor and calibrated moments at every floor. The first three settings share an expected pair-token allowance, their realised generation ratios stay close, and random inclusion together with continuation-length prediction accounts for the small differences in utilization. The .20 setting instead doubles the allowance, as its budget column indicates, and it lowers design variance and extreme advantages while its JEM remains below the default. More observations and better optimization are different coordinates of this experiment, and only the second is what we are after.

At the default floor the predictor mean and variance are \metric{tail_predictor_mean} and \metric{tail_predictor_var}, and after residual correction \metric{tail_adv_mean} and \metric{tail_adv_var}. The predictor's 99th absolute percentile is \metric{tail_predictor_abs_p99} against \metric{tail_adv_p99} for calibrated credit. The top one percent of score-weighted absolute contributions carries \metric{tail_top1_mass}\% of the total mass, which is the number we watch most closely: it says how much of an update rests on a handful of decisions. The .05 design's 99.9th percentile exceeds the .02 realization even though its allowed inverse weight is smaller, and the tail curves preserve that crossing rather than smoothing it away.

\begin{figure*}[!t]
\centering
\includegraphics[width=.49\textwidth]{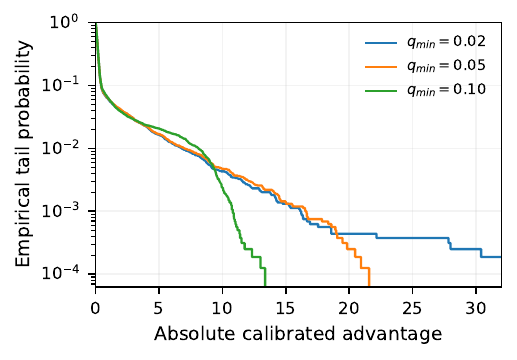}\hfill
\includegraphics[width=.49\textwidth]{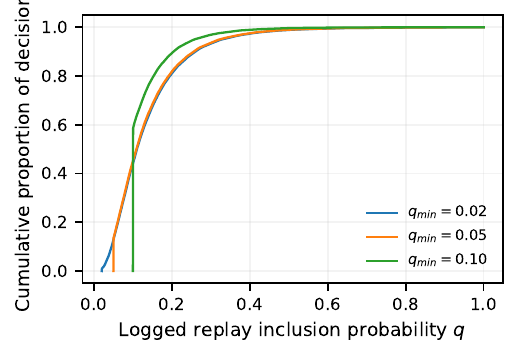}
\caption{Left: absolute calibrated-advantage survival distributions with the full upper tail retained. Right: empirical inclusion-probability distributions on the common 16,000-candidate bank. Floors .02, .05, and .10 share an expected pair-token allowance; the .20 setting uses the expanded allowance of Table~\ref{tab:budget}.}
\label{fig:tails}
\end{figure*}

\begin{table*}[!t]
\centering\tabsetup
\small
\begin{tabular}{rrrrrrr}
\toprule
Refresh & Step $\times$ & Mean $|\delta|$ & 95\% CI & P95 & KL & JEM\\
\midrule
1 & 0.5 & 0.0050 & [0.0046, 0.0053] & 0.0183 & 0.0008 & $37.32\,\pm\,1.28$ \\
1 & 1.0 & 0.0099 & [0.0092, 0.0106] & 0.0362 & 0.0033 & $36.82\,\pm\,1.17$ \\
1 & 2.0 & 0.0196 & [0.0184, 0.0210] & 0.0696 & 0.0132 & $36.40\,\pm\,1.25$ \\
2 & 0.5 & 0.0079 & [0.0074, 0.0085] & 0.0290 & 0.0021 & $36.80\,\pm\,1.33$ \\
2 & 1.0 & 0.0158 & [0.0147, 0.0169] & 0.0567 & 0.0085 & $36.31\,\pm\,1.30$ \\
2 & 2.0 & 0.0310 & [0.0290, 0.0332] & 0.1089 & 0.0337 & $36.37\,\pm\,1.21$ \\
4 & 0.5 & 0.0127 & [0.0119, 0.0135] & 0.0459 & 0.0055 & $36.04\,\pm\,1.43$ \\
4 & 1.0 & 0.0250 & [0.0234, 0.0268] & 0.0875 & 0.0217 & $36.72\,\pm\,1.24$ \\
4 & 2.0 & 0.0479 & [0.0448, 0.0513] & 0.1625 & 0.0843 & $36.56\,\pm\,1.70$ \\
8 & 0.5 & 0.0202 & [0.0188, 0.0215] & 0.0710 & 0.0140 & $36.54\,\pm\,0.82$ \\
8 & 1.0 & 0.0392 & [0.0366, 0.0418] & 0.1345 & 0.0549 & $36.11\,\pm\,1.28$ \\
8 & 2.0 & 0.0710 & [0.0667, 0.0755] & 0.2334 & 0.2042 & $36.38\,\pm\,0.97$ \\
\bottomrule
\end{tabular}
\caption{Continuation-policy refresh and step-size sweep. Refresh is measured in actor updates. $\delta=\Delta_{\mathrm{new}}-\Delta_{\mu}$ holds the reference distribution fixed. Each setting uses 1,200 paired prefixes and 3,000 prefix bootstrap replicates. JEM is the eight-seed MuSiQue endpoint.}
\label{tab:drift}
\end{table*}

\begin{table}[H]
\centering\tabsetup
\small
\begin{tabular}{lrrr}
\toprule
Action & Prefixes & Mean $|\delta|$ & P95\\
\midrule
FINAL & 199 & 0.0055 & 0.0171 \\
READ & 331 & 0.0072 & 0.0273 \\
SEARCH & 394 & 0.0138 & 0.0470 \\
VERIFY & 276 & 0.0108 & 0.0356 \\
\bottomrule
\end{tabular}
\caption{Action-level drift at the default one-update refresh and unit step multiplier. The four strata partition the same 1,200 audited prefixes.}
\label{tab:drift-actions}
\end{table}

A smaller actor step reduces continuation drift, and longer refresh intervals increase it at the same step scale (Table~\ref{tab:drift}). The endpoint ordering is less regular, because a smaller update can help one configuration while excessive staleness weakens the credit the next update relies on. The search stratum has the widest default drift distribution, since alternative retrieval choices change which evidence is reachable afterwards; reading and finalization move less on the common prefix bank.
\begin{figure}[H]
\centering\includegraphics[width=\columnwidth]{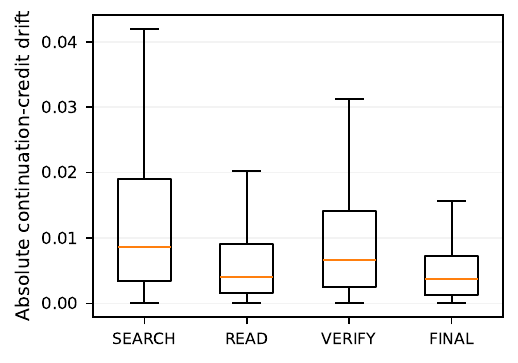}
\caption{Absolute credit drift by action type at one-update refresh and unit step. Boxes are quartiles with the median, whiskers extend to 1.5 interquartile ranges, and outliers are omitted. The same 1,200 prefixes underlie Table~\ref{tab:drift-actions}.}
\label{fig:drift-actions}
\end{figure}

\begin{table*}[!t]
\centering\tabsetup
\small
\begin{tabular}{lrrrrrr}
\toprule
Task & FLOPs & CM-RL & AM-CF & C3 & Uniform & PCCC-RL\\
\midrule
HotpotQA & 0.8 & $49.58\,\pm\,2.03$ & $51.80\,\pm\,1.95$ & $52.17\,\pm\,2.02$ & $51.89\,\pm\,1.37$ & $52.71\,\pm\,1.58$ \\
HotpotQA & 1.6 & $52.25\,\pm\,2.17$ & $52.82\,\pm\,1.39$ & $55.98\,\pm\,1.36$ & $55.60\,\pm\,1.17$ & $57.07\,\pm\,1.13$ \\
HotpotQA & 2.6 & $52.56\,\pm\,1.73$ & $54.56\,\pm\,1.45$ & $54.79\,\pm\,2.28$ & $57.96\,\pm\,2.54$ & $58.50\,\pm\,2.56$ \\
MuSiQue & 0.8 & $28.85\,\pm\,2.90$ & $30.50\,\pm\,2.49$ & $32.37\,\pm\,1.60$ & $31.90\,\pm\,1.93$ & $32.53\,\pm\,1.80$ \\
MuSiQue & 1.6 & $31.26\,\pm\,1.26$ & $33.53\,\pm\,1.75$ & $35.94\,\pm\,2.38$ & $36.27\,\pm\,1.52$ & $36.82\,\pm\,1.24$ \\
MuSiQue & 2.6 & $32.54\,\pm\,1.83$ & $34.92\,\pm\,3.58$ & $37.55\,\pm\,3.56$ & $39.47\,\pm\,3.29$ & $36.81\,\pm\,3.64$ \\
\bottomrule
\end{tabular}
\caption{JEM at common resource locations ($10^{19}$ FLOPs). Values linearly interpolate each seed between its two adjacent saved checkpoints, then average over seeds. The raw checkpoint locations and their dispersion appear in Figure~\ref{fig:frontier}. The six primary tests use un-interpolated endpoints.}
\label{tab:resource-grid}
\end{table*}

\begin{figure*}[!t]
\begin{minipage}[t]{.49\textwidth}
\centering\includegraphics[width=\linewidth]{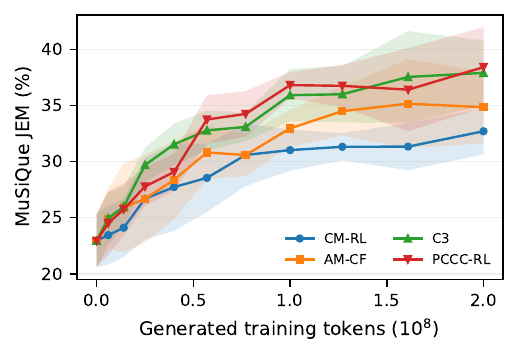}
\caption{MuSiQue learning trajectories. The $10^8$-token checkpoint is the endpoint of Table~\ref{tab:main}; lines are eight-seed means and bands are $\pm$ one sample SD.}
\label{fig:learning}
\end{minipage}\hfill
\begin{minipage}[t]{.49\textwidth}
\centering\includegraphics[width=\linewidth]{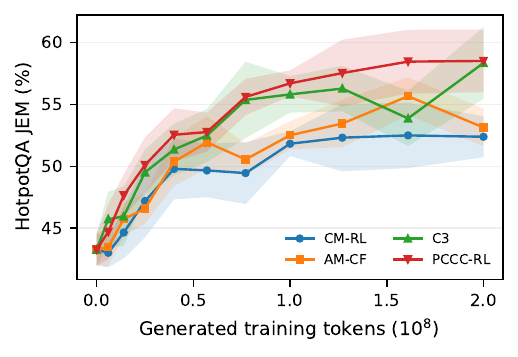}
\caption{HotpotQA learning trajectories from the shared warm start, with eight-seed means and bands. The primary-token endpoint reuses the Table~\ref{tab:main} outcomes exactly.}
\label{fig:learning-hp}
\end{minipage}
\end{figure*}

\subsection{Responses to changed facts and their failures}
The integer counts in Table~\ref{tab:language} recover every displayed family mean. Preservation under irrelevant insertion is easier than response to a changed premise, which is the expected asymmetry: adding inert text leaves the answer supported, while changing a premise forces the agent to abandon an answer it has already found. Relation and value edits benefit from updating the relevant citation together with the answer. Scope remains the least accurate response category and retains a small reversal against C3.

\subsubsection{Controlled case analysis}
\label{app:cases}
Section~\ref{sec:language-results} presents the deletion case. The three below show a relation change, a quantifier failure, and a shifted time boundary, each exposing the edited proposition, the required response, and the corresponding behaviour. Bracketed identifiers refer to the displayed evidence rather than to hidden evaluator fields, and the passages use self-contained local entities so that the gold response follows from the text alone.

\begin{tcolorbox}[enhanced,breakable,colback=TableStripe,colbacktitle=TableStripe,coltitle=black,frame hidden,borderline west={2.5pt}{0pt}{FlowGreen},boxrule=0pt,sharp corners,title={Case B: relation replacement},title after break={Case B: relation replacement (continued)},fonttitle=\small\bfseries]
\small
\textbf{Question.} Who founded Meridian Observatory?\\
\textbf{Original evidence.} [B1] Nila Voss founded Meridian Observatory. [B2] Tomas Reed became its director in 2012.\\
\textbf{Edit.} [B1] Nila Voss \emph{funded} Meridian Observatory. [B2] Tomas Reed \emph{founded} the observatory and became its director in 2012.\\
\textbf{Gold after edit.} Tomas Reed [B2].\\
\textbf{AM-CF.} ``Nila Voss [B1].''\\
\textbf{PCCC-RL.} ``Tomas Reed [B2].''\\
\textbf{Decision trace.} A near-lexical replacement changes the founder relation while preserving both people and the institution. PCCC-RL changes entity and citation together, whereas AM-CF keeps the entity it had already associated with the question.
\end{tcolorbox}

\begin{tcolorbox}[enhanced,breakable,colback=TableStripe,colbacktitle=TableStripe,coltitle=black,frame hidden,borderline west={2.5pt}{0pt}{FlowGreen},boxrule=0pt,sharp corners,title={Case C: a quantifier-scope failure},fonttitle=\small\bfseries]
\small
\textbf{Question.} Did none of the six sensors pass calibration?\\
\textbf{Original evidence.} [C1] None of the six sensors passed calibration.\\
\textbf{Edit.} [C1] Not all of the six sensors passed calibration.\\
\textbf{Gold after edit.} The statement establishes that at least one sensor failed. It does not determine whether none passed.\\
\textbf{AM-CF.} ``Yes. None passed [C1].''\\
\textbf{PCCC-RL.} ``No. Some sensors passed [C1].''\\
\textbf{Decision trace.} Both edited responses fail. AM-CF retains universal failure, while PCCC-RL overcorrects to an existential success that the edited sentence does not entail. The error sits in the scope relation rather than in retrieval or in the cited passage.
\end{tcolorbox}

\begin{tcolorbox}[enhanced,breakable,colback=TableStripe,colbacktitle=TableStripe,coltitle=black,frame hidden,borderline west={2.5pt}{0pt}{FlowGreen},boxrule=0pt,sharp corners,title={Case D: temporal-boundary update},fonttitle=\small\bfseries]
\small
\textbf{Question.} How many clinics had opened by the end of 2019?\\
\textbf{Original evidence.} [D1] Three clinics opened in 2018. [D2] Two more opened in December 2019.\\
\textbf{Edit.} Move the opening in [D2] to January 2020.\\
\textbf{Gold after edit.} Three clinics [D1,D2].\\
\textbf{AM-CF.} ``Five clinics [D1,D2].''\\
\textbf{PCCC-RL.} ``Three clinics; the two later openings were in January 2020 [D1,D2].''\\
\textbf{Decision trace.} The supporting identifiers do not change. The decision turns on a temporal condition inside the cited text, which an identifier-only reward cannot see and which is why we include this family at all.
\end{tcolorbox}

\tablebundle{
\begin{minipage}[t]{.49\textwidth}
\begin{table}[H]
\centering\tabsetup
\small
\begin{tabular}{lrrrrr}
\toprule
Method & \shortstack{FLOPs\\($10^{19}$)} & \shortstack{GPU\\h} & \shortstack{Wall\\h} & \shortstack{Service\\h} & \shortstack{Peak\\GB}\\
\midrule
CM-RL & 1.339 & 119.53 & 16.76 & 1.82 & 59.7 \\
AM-CF & 1.467 & 130.29 & 20.11 & 3.82 & 66.7 \\
C3 & 1.539 & 137.36 & 20.93 & 3.76 & 68.1 \\
Uniform replay & 1.512 & 139.39 & 21.19 & 3.76 & 66.8 \\
PCCC-RL & 1.521 & 138.00 & 21.10 & 3.85 & 67.4 \\
\bottomrule
\end{tabular}
\caption{Equal-token cost ledger: $10^8$ RL-generated tokens, eight accelerators, mean over eight seeds. GPU h sums occupied accelerator time. Wall h includes the serial replay-service component. FLOPs include generation, prefill, model updates, and auxiliary prediction. Service h is displayed separately.}
\label{tab:compute}
\end{table}

\end{minipage}\hfill
\begin{minipage}[t]{.49\textwidth}
\begin{table}[H]
\centering\tabsetup
\setlength{\tabcolsep}{3pt}
\small
\begin{tabular}{lrrrrr}
\toprule
Method & \shortstack{Generation\\$\times$} & \shortstack{FLOPs\\$\times$} & \shortstack{GPU h\\$\times$} & \shortstack{Pair\\M} & \shortstack{Proposal\\M}\\
\midrule
\shortstack[l]{Terminal\\RL} & 1.000 & 1.000 & 1.000 & 0.00 & 0.00 \\
\shortstack[l]{Uniform\\replay} & 2.005 & 2.182 & 2.325 & 43.27 & 6.98 \\
\shortstack[l]{Adaptive\\replay} & 1.884 & 2.030 & 2.227 & 37.66 & 6.55 \\
\shortstack[l]{Four\\pairs} & 4.133 & 4.351 & 4.659 & 150.23 & 6.42 \\
\bottomrule
\end{tabular}
\caption{Fixed-factual-update overhead. Every seed generates $5\times10^7$ factual tokens. Ratios normalize by the terminal-RL mean. Eight seeds are used per row. Four-pair replay shares a reference proposal pool across its repeated continuations.}
\label{tab:fixed-cost}
\end{table}

\end{minipage}
}

\subsection{Citation support, answer coverage, and human review}
\begin{table}[H]
\centering\tabsetup
\setlength{\tabcolsep}{2pt}
\small
\begin{tabular}{lrrrr}
\toprule
Metric & CM-RL & AM-CF & C3 & PCCC-RL \\
\midrule
STR-EM & \shortstack{41.31\\$\pm$1.63} & \shortstack{42.92\\$\pm$0.70} & \shortstack{43.81\\$\pm$1.54} & \shortstack{43.51\\$\pm$2.06} \\
Cit. P & 75.02 & 78.60 & 81.19 & 83.40 \\
Cit. R & 69.05 & 73.47 & 75.49 & 73.51 \\
Tokens & 153.81 & 159.73 & 165.48 & 150.91 \\
Citations & 5.39 & 5.42 & 5.44 & 5.39 \\
\shortstack{Omitted\\facets} & 2.32 & 2.25 & 2.22 & 2.23 \\
\bottomrule
\end{tabular}
\caption{ALCE-ASQA transfer on 948 fixed items, eight seeds. STR-EM, citation precision, and citation recall are macro percentages over answers. Output tokens, citations, and omitted short-answer facets are per-answer means. The same frozen citation checker and passage set are used throughout.}
\label{tab:citation}
\end{table}

\begin{table}[H]
\centering\tabsetup
\small
\begin{tabular}{lrrrr}
\toprule
Method & \shortstack{Fleiss\\$\kappa$} & \shortstack{95\%\\CI} & \shortstack{Pairwise\\(\%)} & \shortstack{Unanimous\\(\%)}\\
\midrule
CM-RL & 0.735 & [0.662, 0.797] & 93.78 & 90.67 \\
AM-CF & 0.654 & [0.565, 0.731] & 93.89 & 90.83 \\
PCCC-RL & 0.695 & [0.600, 0.775] & 95.11 & 92.67 \\
\bottomrule
\end{tabular}
\caption{Agreement from the three independent, pre-adjudication labels per output. Pairwise agreement and unanimity use different numerators. Bootstrap samples retain all three rater labels for each resampled prompt.}
\label{tab:agreement}
\end{table}

On ASQA, higher precision accompanies shorter outputs and slightly lower facet coverage than C3. Citation counts remain similar across methods, so the precision gain is associated with how well each citation supports its claim rather than with attaching more references to the same answer. The output audit is task-dependent: PCCC-RL has fewer flagged HotpotQA and MuSiQue outputs than AM-CF but more flagged ASQA outputs, and the aggregate incidence averages these matched blocks rather than smoothing them.

Fleiss' $\kappa$ and raw agreement move differently because category prevalence changes across methods, so Table~\ref{tab:agreement} reports pairwise and unanimous agreement together with $\kappa$ and its interval. All values use the original three labels, before majority voting.

\subsection{Sensitivity to unreliable checks and differing environments}
\begin{table}[H]
\centering\tabsetup
\small
\begin{tabular}{rrrrr}
\toprule
Noise (\%) & \shortstack{AM-CF\\JEM} & \shortstack{PCCC\\JEM} & \shortstack{AM\\UCR} & \shortstack{PCCC\\UCR}\\
\midrule
0 & $32.96\,\pm\,1.64$ & $36.82\,\pm\,1.17$ & 8.95 & 6.28 \\
5 & $32.78\,\pm\,1.93$ & $36.37\,\pm\,1.48$ & 9.76 & 7.15 \\
10 & $32.10\,\pm\,1.76$ & $34.95\,\pm\,1.18$ & 10.63 & 8.19 \\
20 & $30.09\,\pm\,2.24$ & $30.21\,\pm\,1.31$ & 13.81 & 10.98 \\
\bottomrule
\end{tabular}
\caption{Training-verifier corruption on MuSiQue (eight seeds). Two-thirds of label flips are false positives. Evaluation labels remain fixed. JEM uses the same registered endpoint in the zero-noise and default rows. UCR is the mean claim-level rate across seeds.}
\label{tab:noise}
\end{table}

\begin{table}[H]
\centering\tabsetup
\small
\begin{tabular}{llrr}
\toprule
Backbone & Seeds & AM-CF & PCCC-RL \\
\midrule
\multicolumn{4}{l}{\textbf{HotpotQA}} \\
Qwen2.5-7B & 8 & $52.52\,\pm\,1.13$ & $56.71\,\pm\,1.05$ \\
Llama-3.1-8B & 3 & $53.87\,\pm\,1.74$ & $57.97\,\pm\,1.40$ \\
Qwen2.5-14B & 3 & $56.77\,\pm\,2.21$ & $61.25\,\pm\,1.98$ \\
\multicolumn{4}{l}{\textbf{MuSiQue}} \\
Qwen2.5-7B & 8 & $32.96\,\pm\,1.64$ & $36.82\,\pm\,1.17$ \\
Llama-3.1-8B & 3 & $33.94\,\pm\,2.34$ & $36.42\,\pm\,1.39$ \\
Qwen2.5-14B & 3 & $38.08\,\pm\,2.84$ & $40.82\,\pm\,2.46$ \\
\bottomrule
\end{tabular}
\caption{Backbone transfer with the reference construction, floor, and replay budget fixed. Values are JEM mean $\pm$ SD. The default backbone row reuses all eight primary seeds. The additional backbones use seeds 11, 23, and 37.}
\label{tab:scales}
\end{table}

\begin{table}[H]
\centering\tabsetup
\setlength{\tabcolsep}{3pt}
\small
\begin{tabular}{lrrr}
\toprule
Method & ALFWorld & ScienceWorld & WebShop\\
\midrule
CM-RL & $71.64\,\pm\,6.38$ & $67.11\,\pm\,2.45$ & $65.40\,\pm\,4.19$ \\
AM-CF & $71.89\,\pm\,2.15$ & $67.89\,\pm\,0.92$ & $63.73\,\pm\,3.74$ \\
C3 & $77.86\,\pm\,3.02$ & $70.79\,\pm\,2.02$ & $66.20\,\pm\,1.06$ \\
PCCC-RL & $74.38\,\pm\,1.88$ & $72.90\,\pm\,1.39$ & $65.60\,\pm\,2.91$ \\
\bottomrule
\end{tabular}
\caption{Secondary environment transfer (three paired seeds). ALFWorld uses 134 unseen tasks and binary success. ScienceWorld uses 300 task--variation instances and normalized native score. WebShop uses 500 goals and binary purchase success. All columns are on a 0--100 scale, with their native metric retained.}
\label{tab:transfer}
\end{table}

Verifier corruption preferentially introduces false positives, and the clean evaluation scorer is held fixed across the sweep. At 20\% noise the large clean-data JEM gap contracts to a small difference, and unsupported-claim rates rise for both methods, which places the counterfactual label source alongside the allocation mechanism in how we explain the result.

The scale comparison reuses the eight-seed 7B endpoints and adds three paired seeds for Llama-3.1-8B and Qwen2.5-14B, holding the task interface and optimization settings fixed. Transfer to ALFWorld, ScienceWorld, and WebShop uses three seeds and each environment's native metric \citep{shridhar2021alfworld,wang2022scienceworld,yao2022webshop}. ALFWorld contributes 134 evaluation tasks, ScienceWorld 300 scored task variants, and WebShop 500 shopping episodes, and the smaller samples yield visibly wider uncertainty than the primary QA results.

\subsection{Training progress and the full compute accounts}
The primary resource ledger sits at a common generated-token allowance, whereas the fixed-factual ledger holds factual work constant. Replay changes both the fraction of tokens spent on useful trajectories and the amount of auxiliary computation, and service time separates accelerator-hours from elapsed training time. We display these quantities independently because collapsing them into a single generation multiplier hides exactly the trade-off the method makes.

The two learning plots start from identical within-seed warm starts and retain intermediate regressions. Resource-grid entries use piecewise-linear interpolation at common FLOP coordinates, with seed variation computed after interpolation, which supplements the un-interpolated checkpoint frontiers and makes the selected operating points easy to compare. Final tables, narrative values, and chart endpoints all come from the same analysis pipeline.

\FloatBarrier


\addcontentsline{toc}{section}{References}
\begin{thebibliography}{25}
\providecommand{\natexlab}[1]{#1}

\bibitem[{Achiam et~al.(2017)Achiam, Held, Tamar, and Abbeel}]{achiam2017cpo}
Joshua Achiam, David Held, Aviv Tamar, and Pieter Abbeel. 2017.
\newblock \href {https://proceedings.mlr.press/v70/achiam17a.html}
  {{Constrained Policy Optimization}}.
\newblock In \emph{Proceedings of the 34th International Conference on Machine
  Learning}, pages 22--31.

\bibitem[{Chen et~al.(2026)Chen, Sun, Wang, Wang, Zhang, Shen, Li, and
  Zhang}]{chen2026c3}
Yanjun Chen, Yirong Sun, Hanlin Wang, Jinghan Wang, Xinming Zhang, Xiaoyu Shen,
  Wenjie Li, and Wei Zhang. 2026.
\newblock \href {https://arxiv.org/abs/2603.06859v4} {{The Trace Is the State:
  Exact Credit Assignment for LLM Agent Teams}}.
\newblock arXiv preprint arXiv:2603.06859, version 4.

\bibitem[{Ding et~al.(2026)Ding, Zhang, Wang, and Zeng}]{ding2026ibpo}
Fei Ding, Yongkang Zhang, Youwei Wang, and Zijian Zeng. 2026.
\newblock \href {https://arxiv.org/abs/2605.16302} {{Reducing Credit Assignment
  Variance via Counterfactual Reasoning Paths}}.
\newblock arXiv preprint arXiv:2605.16302.

\bibitem[{Dror et~al.(2018)Dror, Baumer, Shlomov, and
  Reichart}]{dror2018significance}
Rotem Dror, Gili Baumer, Segev Shlomov, and Roi Reichart. 2018.
\newblock \href {https://aclanthology.org/P18-1128/} {{The Hitchhiker's Guide
  to Testing Statistical Significance in Natural Language Processing}}.
\newblock In \emph{Proceedings of the 56th Annual Meeting of the Association
  for Computational Linguistics (Volume 1: Long Papers)}, pages 1383--1392.

\bibitem[{Dudík et~al.(2011)Dudík, Langford, and Li}]{dudik2011dr}
Miroslav Dudík, John Langford, and Lihong Li. 2011.
\newblock \href {https://arxiv.org/abs/1103.4601} {{Doubly Robust Policy
  Evaluation and Learning}}.
\newblock In \emph{Proceedings of the 28th International Conference on Machine
  Learning}, pages 1097--1104.

\bibitem[{Foerster et~al.(2018)Foerster, Farquhar, Afouras, Nardelli, and
  Whiteson}]{foerster2018coma}
Jakob~N. Foerster, Gregory Farquhar, Triantafyllos Afouras, Nantas Nardelli,
  and Shimon Whiteson. 2018.
\newblock \href {https://arxiv.org/abs/1705.08926} {{Counterfactual Multi-Agent
  Policy Gradients}}.
\newblock In \emph{Proceedings of the AAAI Conference on Artificial
  Intelligence}.

\bibitem[{Gao et~al.(2023)Gao, Yen, Yu, and Chen}]{gao2023alce}
Tianyu Gao, Howard Yen, Jiatong Yu, and Danqi Chen. 2023.
\newblock \href {https://aclanthology.org/2023.emnlp-main.398/} {{Enabling
  Large Language Models to Generate Text with Citations}}.
\newblock In \emph{Proceedings of the 2023 Conference on Empirical Methods in
  Natural Language Processing}, pages 6498--6520.

\bibitem[{Ho et~al.(2020)Ho, Nguyen, Sugawara, and Aizawa}]{ho2020two}
Xanh Ho, Anh-Khoa~Duong Nguyen, Saku Sugawara, and Akiko Aizawa. 2020.
\newblock \href {https://aclanthology.org/2020.coling-main.580/} {{Constructing
  A Multi-hop QA Dataset for Comprehensive Evaluation of Reasoning Steps}}.
\newblock In \emph{Proceedings of the 28th International Conference on
  Computational Linguistics}, pages 6609--6625.

\bibitem[{Holm(1979)}]{holm1979}
Sture Holm. 1979.
\newblock \href {https://www.jstor.org/stable/4615733} {{A Simple Sequentially
  Rejective Multiple Test Procedure}}.
\newblock \emph{Scandinavian Journal of Statistics, 6(2):65--70}.

\bibitem[{Huang et~al.(2024)Huang, Wu, Hu, and Wang}]{huang2024citations}
Chengyu Huang, Zeqiu Wu, Yushi Hu, and Wenya Wang. 2024.
\newblock \href {https://aclanthology.org/2024.acl-long.161/} {{Training
  Language Models to Generate Text with Citations via Fine-grained Rewards}}.
\newblock In \emph{Proceedings of the 62nd Annual Meeting of the Association
  for Computational Linguistics (Volume 1: Long Papers)}, pages 2926--2949.

\bibitem[{Huang and Jiang(2020)}]{huang2020drpg}
Jiawei Huang and Nan Jiang. 2020.
\newblock \href {https://proceedings.mlr.press/v119/huang20b.html} {{From
  Importance Sampling to Doubly Robust Policy Gradient}}.
\newblock In \emph{Proceedings of the 37th International Conference on Machine
  Learning}, pages 4434--4443.

\bibitem[{Jiang and Li(2016)}]{jiang2016dr}
Nan Jiang and Lihong Li. 2016.
\newblock \href {https://proceedings.mlr.press/v48/jiang16.html} {{Doubly
  Robust Off-policy Value Evaluation for Reinforcement Learning}}.
\newblock In \emph{Proceedings of the 33rd International Conference on Machine
  Learning}, pages 652--661.

\bibitem[{Jin et~al.(2025)Jin, Zeng, Yue, Yoon, Arik, Wang, Zamani, and
  Han}]{jin2025search}
Bowen Jin, Hansi Zeng, Zhenrui Yue, Jinsung Yoon, Sercan Arik, Dong Wang, Hamed
  Zamani, and Jiawei Han. 2025.
\newblock \href {https://arxiv.org/abs/2503.09516} {{Search-R1: Training LLMs
  to Reason and Leverage Search Engines with Reinforcement Learning}}.
\newblock arXiv preprint arXiv:2503.09516.

\bibitem[{Lewis et~al.(2020)Lewis, Perez, Piktus, Petroni, Karpukhin, Goyal,
  Küttler, Lewis, Yih, Rocktäschel, Riedel, and Kiela}]{lewis2020rag}
Patrick Lewis, Ethan Perez, Aleksandra Piktus, Fabio Petroni, Vladimir
  Karpukhin, Naman Goyal, Heinrich Küttler, Mike Lewis, Wen-tau Yih, Tim
  Rocktäschel, Sebastian Riedel, and Douwe Kiela. 2020.
\newblock \href {https://arxiv.org/abs/2005.11401} {{Retrieval-Augmented
  Generation for Knowledge-Intensive NLP Tasks}}.
\newblock In \emph{Advances in Neural Information Processing Systems, 33},
  pages 9459--9474.

\bibitem[{Owen(2007)}]{owen2007}
Art~B. Owen. 2007.
\newblock \href {https://arxiv.org/abs/0712.1111} {{The Pigeonhole Bootstrap}}.
\newblock \emph{The Annals of Applied Statistics, 1(2):386--411}.

\bibitem[{{Qwen Team}(2025)}]{yang2025qwen}
{Qwen Team}. 2025.
\newblock \href {https://arxiv.org/abs/2412.15115v2} {{Qwen2.5 Technical
  Report}}.
\newblock arXiv preprint arXiv:2412.15115, version 2.

\bibitem[{Schulman et~al.(2017)Schulman, Wolski, Dhariwal, Radford, and
  Klimov}]{schulman2017ppo}
John Schulman, Filip Wolski, Prafulla Dhariwal, Alec Radford, and Oleg Klimov.
  2017.
\newblock \href {https://arxiv.org/abs/1707.06347} {{Proximal Policy
  Optimization Algorithms}}.
\newblock arXiv preprint arXiv:1707.06347.

\bibitem[{Shao et~al.(2024)Shao, Wang, Zhu, Xu, Song, Bi, Zhang, Zhang, Li, Wu,
  and Guo}]{shao2024deepseekmath}
Zhihong Shao, Peiyi Wang, Qihao Zhu, Runxin Xu, Junxiao Song, Xiao Bi, Haowei
  Zhang, Mingchuan Zhang, Y.~K. Li, Y.~Wu, and Daya Guo. 2024.
\newblock \href {https://arxiv.org/abs/2402.03300} {{DeepSeekMath: Pushing the
  Limits of Mathematical Reasoning in Open Language Models}}.
\newblock arXiv preprint arXiv:2402.03300.

\bibitem[{Shridhar et~al.(2021)Shridhar, Yuan, Côté, Bisk, Trischler, and
  Hausknecht}]{shridhar2021alfworld}
Mohit Shridhar, Xingdi Yuan, Marc-Alexandre Côté, Yonatan Bisk, Adam
  Trischler, and Matthew Hausknecht. 2021.
\newblock \href {https://arxiv.org/abs/2010.03768} {{ALFWorld: Aligning Text
  and Embodied Environments for Interactive Learning}}.
\newblock In \emph{International Conference on Learning Representations}.

\bibitem[{Tran et~al.(2026)Tran, Yao, and Yu}]{tran2026tempo}
Hieu Tran, Zonghai Yao, and Hong Yu. 2026.
\newblock \href {https://aclanthology.org/2026.findings-acl.524/} {{Exploiting
  Tree Structure for Credit Assignment in Reinforcement Learning with Large
  Language Models}}.
\newblock In \emph{Findings of the Association for Computational Linguistics:
  ACL 2026}, pages 10795--10810.

\bibitem[{Trivedi et~al.(2022)Trivedi, Balasubramanian, Khot, and
  Sabharwal}]{trivedi2022musique}
Harsh Trivedi, Niranjan Balasubramanian, Tushar Khot, and Ashish Sabharwal.
  2022.
\newblock \href {https://aclanthology.org/2022.tacl-1.31/} {{MuSiQue: Multihop
  Questions via Single-hop Question Composition}}.
\newblock \emph{Transactions of the Association for Computational Linguistics,
  10:539--554}.

\bibitem[{Wang et~al.(2022)Wang, Jansen, Côté, and
  Ammanabrolu}]{wang2022scienceworld}
Ruoyao Wang, Peter Jansen, Marc-Alexandre Côté, and Prithviraj Ammanabrolu.
  2022.
\newblock \href {https://aclanthology.org/2022.emnlp-main.775/} {{ScienceWorld:
  Is your Agent Smarter than a 5th Grader?}}
\newblock In \emph{Proceedings of the 2022 Conference on Empirical Methods in
  Natural Language Processing}, pages 11279--11298.

\bibitem[{Yang et~al.(2018)Yang, Qi, Zhang, Bengio, Cohen, Salakhutdinov, and
  Manning}]{yang2018hotpot}
Zhilin Yang, Peng Qi, Saizheng Zhang, Yoshua Bengio, William~W. Cohen, Ruslan
  Salakhutdinov, and Christopher~D. Manning. 2018.
\newblock \href {https://aclanthology.org/D18-1259/} {{HotpotQA: A Dataset for
  Diverse, Explainable Multi-hop Question Answering}}.
\newblock In \emph{Proceedings of the 2018 Conference on Empirical Methods in
  Natural Language Processing}, pages 2369--2380.

\bibitem[{Yao et~al.(2022)Yao, Chen, Yang, and Narasimhan}]{yao2022webshop}
Shunyu Yao, Howard Chen, John Yang, and Karthik Narasimhan. 2022.
\newblock \href {https://arxiv.org/abs/2207.01206} {{WebShop: Towards Scalable
  Real-World Web Interaction with Grounded Language Agents}}.
\newblock In \emph{Advances in Neural Information Processing Systems, 35},
  pages 20744--20757.

\bibitem[{Yao et~al.(2023)Yao, Zhao, Yu, Du, Shafran, Narasimhan, and
  Cao}]{yao2023react}
Shunyu Yao, Jeffrey Zhao, Dian Yu, Nan Du, Izhak Shafran, Karthik Narasimhan,
  and Yuan Cao. 2023.
\newblock \href {https://arxiv.org/abs/2210.03629} {{ReAct: Synergizing
  Reasoning and Acting in Language Models}}.
\newblock In \emph{International Conference on Learning Representations}.

\end{thebibliography}
\end{document}